\documentclass[lettersize,journal]{IEEEtran}
\usepackage{amsmath,amsfonts}
\usepackage[ruled,linesnumbered,lined]{algorithm2e}
\usepackage{algorithmic}
\usepackage{array}
\usepackage[caption=false,font=normalsize,labelfont=sf,textfont=sf]{subfig}
\usepackage{textcomp}
\usepackage{stfloats}
\usepackage{url}
\usepackage{verbatim}
\usepackage{graphicx}
\usepackage{cite}
\usepackage{pifont}
\usepackage{makecell}
\usepackage{bm}
\usepackage{bbm}
\usepackage[english]{babel}
\usepackage{amsthm}
\usepackage{color}
\usepackage{soul}

\newcommand{\isEquivTo}[1]{\underset{#1}{\sim}}
\newtheorem{theorem}{Theorem}

\newtheorem{assumption}{Assumption}

\newtheorem{definition}{Definition}
\begin{document}

\title{Multi-Fidelity Bayesian Optimization With Across-Task Transferable Max-Value \\ Entropy Search}

\author{Yunchuan Zhang, \IEEEmembership{Member,~IEEE}, Sangwoo Park, \IEEEmembership{Member,~IEEE}, \\and~Osvaldo Simeone, \IEEEmembership{Fellow,~IEEE}
\thanks{ 
The authors are with the King’s Communications, Learning and Information Processing (KCLIP) lab within the Centre for Intelligent Information Processing Systems (CIIPS), Department of Engineering, King’s College London, London, WC2R 2LS, UK. (email:\{yunchuan.zhang, sangwoo.park, osvaldo.simeone\}@kcl.ac.uk). The work of O. Simeone was supported by European Union’s Horizon Europe project CENTRIC (101096379), by the Open Fellowships of the EPSRC (EP/W024101/1), and by the EPSRC project (EP/X011852/1).
}
}



\maketitle

\begin{abstract}
In many applications, ranging from logistics to engineering, a designer is faced with a sequence of optimization tasks for which the objectives are in the form of black-box functions that are costly to evaluate. Furthermore, higher-fidelity evaluations of the optimization objectives often entail a larger cost. Existing multi-fidelity black-box optimization strategies select candidate solutions and fidelity levels with the goal of maximizing the information about the optimal value or the optimal solution for the current task. Assuming that successive optimization tasks are related, this paper introduces a novel information-theoretic acquisition function that balances the need to acquire information about the current task with the goal of collecting information transferable to future tasks. The proposed method transfers across tasks distributions over parameters of a Gaussian process surrogate model by implementing particle-based variational Bayesian updates. Theoretical insights based on the analysis of the expected regret substantiate the benefits of acquiring transferable knowledge across tasks. Furthermore, experimental results across synthetic and real-world examples reveal that the proposed acquisition strategy that caters to future tasks can significantly improve the optimization efficiency as soon as a sufficient number of tasks is processed.
\end{abstract}

\begin{IEEEkeywords}
Bayesian optimization, multi-fidelity simulation, entropy search, knowledge transfer
\end{IEEEkeywords}

\section{Introduction}\label{sec: intro}
\subsection{Context and Scope}\label{ssec: context and scope}
\IEEEPARstart{N}{umerous} problems in logistics, science, and engineering can be formulated as \emph{black-box optimization} tasks, in which the objective is costly to evaluate. Examples include hyperparameter optimization for machine learning \cite{zhang2021convolutional}, malware detection \cite{demetrio2021functionality}, antenna design \cite{zhou2020trust}, text to speech adaptation \cite{moss2020boffin}, and resource allocation in wireless communication systems \cite{Maggi2021bogp,zhang2023bayesian,johnston2023curriculum}. To mitigate the problem of evaluating a costly objective function for each candidate solutions, the designer may have access to cheaper \emph{approximations} of the optimization target. For example, the designer may be able to \emph{simulate} a physical system using a \emph{digital twin} that  offers a controllable trade-off between \emph{cost} and \emph{fidelity} of the approximation \cite{hoydis2023sionna,ruah2023bayesian,ruah2023calibrating}. As shown in Fig. 1, higher-fidelity evaluations of the objective functions generally entail a larger cost, and the main challenge for the designer is to select a sequence of candidate solutions and fidelity levels that obtains the best solution within the available cost budget.

As a concrete example, consider the problem of optimizing the time spent by  patients in a hospital's emergency department \cite{chen2022multi}. The hospital may try different allocations of medical personnel  by carrying out expensive real-world trials. Alternatively, one may adopt a simulator of patients' hospitalization experiences, with different accuracy levels requiring a larger computing cost in terms of time and energy. 

\begin{figure*}[t]

  \centering
  \centerline{\includegraphics[scale=0.45]{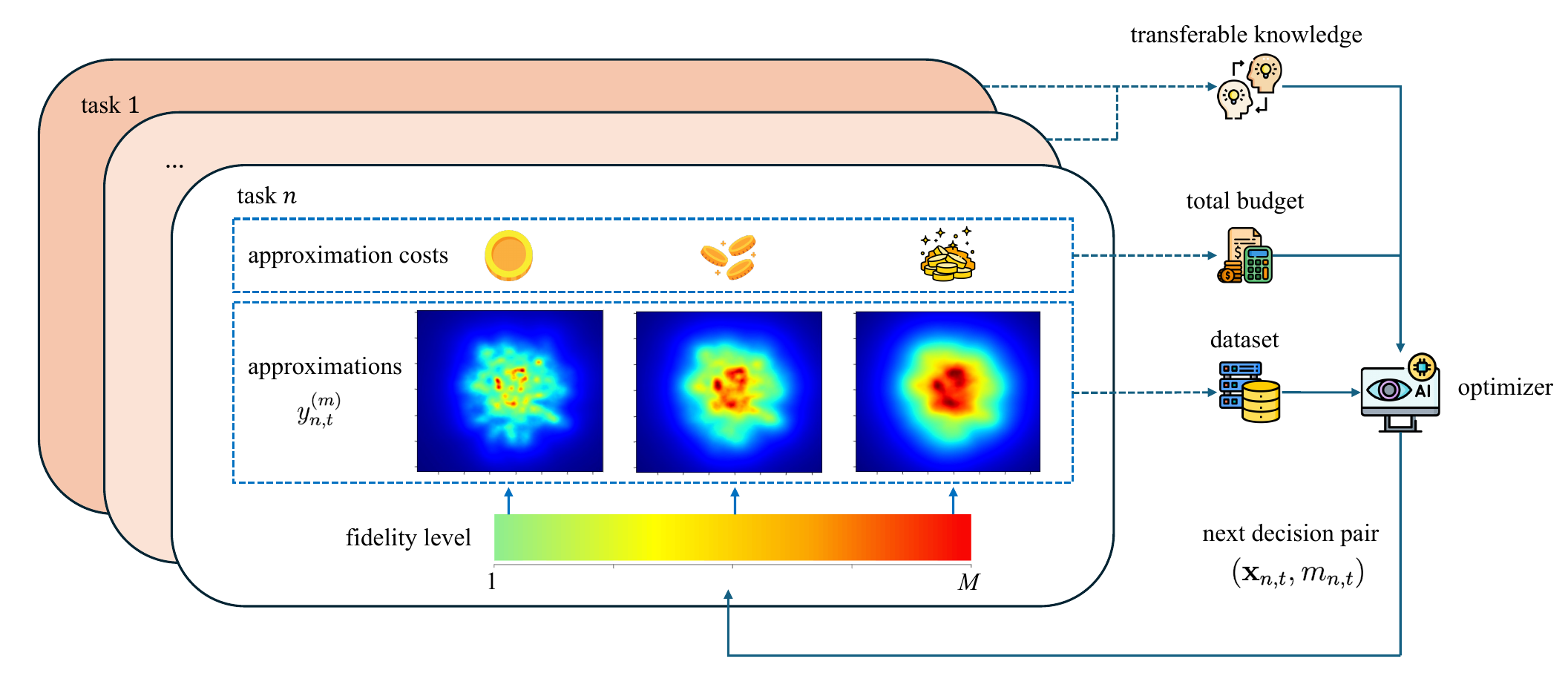}}
  \caption{This paper studies a sequential multi-task optimization setting with multi-fidelity approximations of expensive-to-evaluate black-box objective functions. For any current task $n$, over time index $t=1,2,...,T_n$, the optimizer selects a pair of query point $\mathbf{x}_{n,t}$ and fidelity level $m_{n,t}$, requiring an approximation cost $\lambda^{(m)}$. As a result, the optimizer receives noisy feedback $y_{n,t}^{(m)}$ about target objective value $f_n(\mathbf{x}_{n,t})$. We wish to approach the global optimal solution $\mathbf{x}_n^*$ of objective $f_n(\mathbf{x})$ while abiding by a total simulation cost budget $\Lambda$.}
  \label{fig: intro flow}
\end{figure*}

As also illustrated in Fig. \ref{fig: intro flow}, in many applications, the designer is faced with a \emph{sequence} of black-box optimization tasks for which the objectives are distinct, but related. For instance, one may need to tune the hyperparameters of  neural network models for different learning tasks over time; address the optimal allocation of personnel in a  hospital in different periods of the year; or optimize resource allocation in a wireless system as the users' demands change over time. As detailed in the next section, existing multi-fidelity black-box optimization strategies select
candidate solutions and fidelity levels with the goal of maximizing the information accrued about the
optimal value or solution for the \emph{current} task. 

This paper introduces  a novel information-theoretic selection process for the next candidate solution and fidelity level that balances the need to acquire information about the \emph{current} task with the goal of collecting information transferable to \emph{future} tasks. The proposed method introduces \emph{shared} latent variables across tasks. These variables are  transferred across successive tasks by adopting a Bayesian formalism whereby the posterior distribution at the end of the current task is adopted as prior for the next task.

\subsection{Related Work}\label{ssec: related work}

\emph{Bayesian optimization} (BO) is a popular framework for black-box optimization problems. BO relies on a \emph{surrogate model}, typically a \emph{Gaussian process} (GP) \cite{Rasmussen2004}, which encodes  the current belief of the optimizer about the objective function,  and an \emph{acquisition function} that  selects the next candidate solution based on the surrogate model \cite{jones1998efficient,frazier2008knowledge,wang2022tight,wang2017max,srinivas2012tit}. BO has been extended to address multi-fidelity -- also known as multi-task or multi-information source -- settings  \cite{moss2021gibbon,swersky2013multi,poloczek2017multi}. Via \emph{multi-fidelity BO} (MFBO), information collected at lower fidelity levels can be useful to accelerate the optimization process when viewed as a function of the overall cost budget for evaluating the objective function. 

Prior works developed MFBO by building on  standard BO acquisition functions, including expected improvement (EI) in \cite{lam2015multifidelity}, upper confidence bound (UCB) in \cite{kandasamy2017multi}, and knowledge gradient (KG) in \cite{poloczek2017multi}.  EI-based MFBO does not account for the level of uncertainty in the surrogate model. This issue is addressed by UCB-based approaches, which, however, require a carefully selected  parameter to balance exploitation and exploration. Finally, although KG-based methods achieve efficient global optimization without hyperparameters in the acquisition function, empirical studies in \cite{moss2021gibbon} show that they incur a high computational overhead.

In order to mitigate the limitations of the aforementioned standard acquisition functions, reference \cite{swersky2013multi} introduced an information-theoretic acquisition function based on \emph{entropy search} (ES) \cite{hennig2012entropy}. ES-based MFBO seeks for the next candidate solution by maximizing the \emph{information gain per unit cost}. Accordingly, the ES-based acquisition functions aim to reduce the uncertainty about the global optimum, rather than to improve the current best solution as in EI- and KG-based methods, or to explore the most uncertain regions as in UCB-based approaches. 

Reference \cite{moss2021mumbo}  reduced the computation load of ES-based MFBO via \emph{max-value entropy search} (MES) \cite{wang2017max}; while the work \cite{takeno2020multi} investigated parallel MFBO extensions. Theoretical and empirical comparisons between a light-weight MF-MES framework and other MFBO approaches are carried out in \cite{moss2021gibbon}. As shown recently in \cite{mikkola2023multi}, the robustness of MF-MES can be guaranteed by introducing a novel mechanism of pseudo-observations when the feedback from lower fidelity levels is unreliable.

As illustrated in Fig. 2, the other axis of generalization of BO of interest for this work, namely \emph{transferability} across tasks, has been much less investigated. This line of work relies on the assumption that  sequentially arriving optimization tasks are statistically correlated, such that  knowledge extracted from one task can be transferred to future tasks in the form of an optimized inductive bias encoded into the optimizer \cite{vinyals2016matching}. Existing studies fall into the categories of \emph{lifelong BO}, which leverages previously trained deep neural networks to accelerate the optimizer training process exclusively on the new task \cite{zhang2019lifelong}; and \emph{meta-learned BO}, which learns a well-calibrated prior on the surrogate model given datasets collected from previous tasks \cite{zhang2023bayesian,rothfuss2023meta}. Using these methods, BO is seen to successfully transfer shared information across tasks, providing faster convergence on later tasks. However, these studies are limited to single-fidelity settings. 

\subsection{Main Contributions}\label{ssec: main contri}
Assuming that successive optimization tasks are related, this paper introduces  a novel \emph{information-theoretic} acquisition function that balances the need to acquire information about the current task with the goal of collecting information transferable to future tasks. The proposed method, referred to as \emph{multi-fidelity transferable max-value entropy search} (MFT-MES), includes shared GP model parameters, which are transferred across tasks by implementing particle-based variational Bayesian updates. 

The main contributions are as follows.
\begin{itemize}
    \item We introduce the MFT-MES, a novel black-box optimization scheme tailored for settings in which the designer is faced with a sequence of related optimization tasks. MFT-MES builds on MF-MES \cite{moss2021gibbon} by selecting candidate solutions and fidelity levels that maximize the information gain per unit cost. The information gain in MFT-MES accounts not only for the information about the optimal value of the current objective, as in MF-MES, but also for the information accrued on the shared GP model parameters that can be transferred to future tasks. To this end, MFT-MES models the latent parameters as random quantities whose distributions are updated and transferred across tasks. 
    \item As an efficient implementation of MFT-MES, we propose a particle-based \emph{variational inference} (VI) update strategy for the latent shared parameters by leveraging Stein variational gradient descent (SVGD) \cite{liu2016stein}.
    \item We provide an analysis of the expected regret bound for a stylized version of MFT-MES. The theoretical results substantiate the benefits of exploring the candidate solutions that are likely to provide knowledge transferable across tasks.
    \item We present experimental results across synthetic tasks \cite{dixon1978global} and real-world examples \cite{zhang2023bayesian,li2017fast}. The results reveal that the  provident acquisition strategy implemented by MFT-MES, which caters to future tasks, can significantly improve the optimization efficiency as soon as a sufficient number of tasks is processed.
\end{itemize}

This work was partially presented as a conference paper in  \cite{zhang2024spawc}. The conference version presents a brief presentation of the proposed acquisition strategy, along with limited experiments. In contrast, this paper provides full details on the proposed MFT-MES scheme, along  with a  theoretical analysis of the expected regret bound, as well as with comprehensive experimental results.

\subsection{Organization}

The rest of the paper is organized as follows. Sec. \ref{sec: pre} formulates the sequential multi-task black-box optimization problem, and reviews the MF-GP surrogate model considered in the paper. Sec. \ref{sec: single mfbo} presents the baseline implementation of MF-MES, and illustrates the optimization over surrogate model parameters. The proposed MFT-MES method and the Bayesian update of the shared parameters are introduced in Sec. \ref{sec: k-mfbo}. A theoretical analysis of MFT-MES is provided in Sec. \ref{sec: theorems}. Experimental results on synthetic optimization tasks and real-world applications are provided in Sec. \ref{sec: exp}. Finally, Sec. \ref{sec: conclusion} concludes the paper.

\section{Problem Definition And Preliminaries}\label{sec: pre}
\subsection{Sequential Multi-Task Black-Box Optimization}\label{ssec: problem def}
We consider a setting in which optimization tasks, defined on a common input space $\mathcal{X}\subseteq\mathbbm{R}^d$, are addressed sequentially. Each $n$-th task, with $n=1,2,...$, consists of the optimization of a \emph{black-box expensive-to-evaluate} objective function $f_n(\mathbf{x})$. Examples include the optimization of hyperparameters for machine learning models and experimental design \cite{malkomes2016bayesian}. The objective functions $f_n(\mathbf{x})$ are assumed to be drawn according to a parametric stochastic process $\mathcal{P}_{\boldsymbol{\theta}}(f(\mathbf{x}))$ in an independent identical distributed (i.i.d.) manner, i.e.,
\begin{align} 
    f_n(\mathbf{x})\isEquivTo{\text{i.i.d.}} \mathcal{P}_{\boldsymbol{\theta}}(f(\mathbf{x})) ~\,\text{for}~\, n=1,2,...\label{eq: sample f_n}
\end{align}
Furthermore, the parameter vector $\boldsymbol{\theta}$ identifying the stochastic process $\mathcal{P}_{\boldsymbol{\theta}}(f(\mathbf{x}))$ is unknown, and it is assigned a prior distribution $p(\boldsymbol{\theta})$, i.e.,
\begin{align}
    \boldsymbol{\theta}\sim p(\boldsymbol{\theta}). \label{eq: theta prior}
\end{align}
Note that, by assumptions \eqref{eq: sample f_n} and \eqref{eq: theta prior}, the objective functions $f_1(\mathbf{x}), f_2(\mathbf{x}),...$ are not independent unless one conditions on parameter vector $\boldsymbol{\theta}$. In fact, having information on any function $f_n(\mathbf{x})$ reduces uncertainty on the parameter vector $\boldsymbol{\theta}$, thus providing information about other functions $f_{n'}(\mathbf{x})$ with $n'\neq n$.

For any current $n$-th task, the goal is to obtain an approximation of the optimal solution
\begin{align}
    \mathbf{x}_n^* = \arg\max_{\mathbf{x}\in\mathcal{X}}f_n(\mathbf{x}) \label{eq: opt goal}
\end{align}
with the minimal number of evaluations of function $f_n(\mathbf{x})$. To this end, this paper investigates the idea of selecting candidate solution $\mathbf{x}$ to query the current function $f_n(\mathbf{x})$ not only with the aim of approaching the solution in \eqref{eq: opt goal} for task $n$, but also to extract information about the common parameters $\boldsymbol{\theta}$ that may be useful for future optimization tasks $n'>n$. This way, while convergence to a solution \eqref{eq: opt goal} may be slower for the current task $n$, future tasks may benefit from the acquired knowledge about parameters $\boldsymbol{\theta}$ to speed up convergence.

In order to account for the \emph{cost} of accessing the objective function $f_n(\mathbf{x})$, we follow the \emph{multi-fidelity} formulation, whereby evaluating function $f_n(\mathbf{x})$ at some querying point $\mathbf{x}$ with fidelity level $m$ entails a cost $\lambda^{(m)}>0$ \cite{forrester2007multi}. Different fidelity levels may correspond to training processes with varying number of iterations for hyperparameters optimization, or to simulations of a physical process with varying levels of accuracy for experimental design. 

\begin{figure}[t]

  \centering
  \centerline{\includegraphics[scale=0.46]{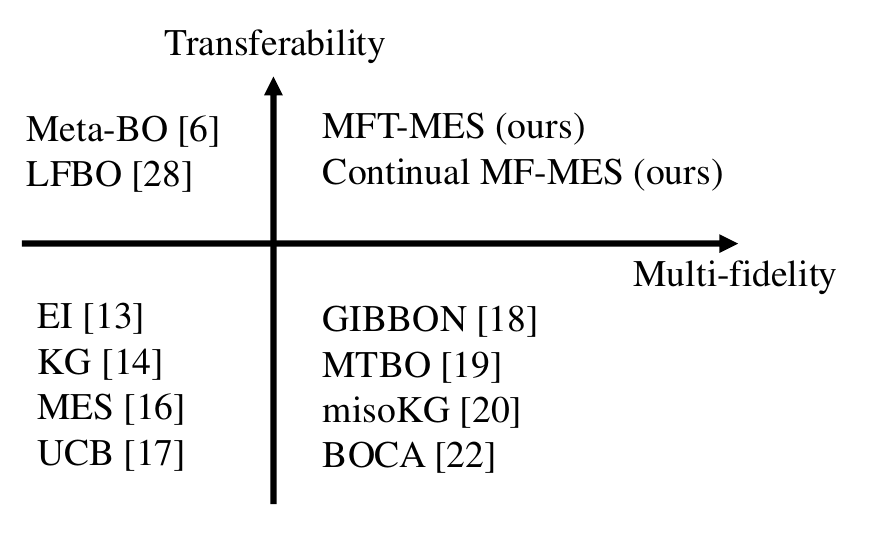}}
  \caption{Comparison between selected prior works and the proposed  MFT-MES method.}\label{fig: related}
%
\end{figure}

There are $M$ fidelity levels, listed from lower fidelity, $m=1$, to highest fidelity, $m=M$, which are collected in set $\mathcal{M}=\{1,2,...,M\}$. The function approximating objective $f_n(\mathbf{x})$ at the $m$-th fidelity level is denoted as $f_n^{(m)}(\mathbf{x})$. The costs are ordered from lowest fidelity to highest fidelity as
\begin{align}
    \lambda^{(1)}\leq\lambda^{(2)}\leq ...\leq\lambda^{(M)}, \label{eq: cost list}
\end{align}
and the highest-fidelity approximation coincides with the true objective function, i.e.,
\begin{align}
    f^{(M)}_n(\mathbf{x})=f_n(\mathbf{x}).\label{eq: highest fidelity}
\end{align}

For each task $n$, the optimizer queries the objective function $f_n(\mathbf{x})$ during $T_n$ \emph{rounds}, choosing at each round $t=1,...,T_n$, an input $\mathbf{x}_{n,t}$ and a fidelity level $m_{n,t}$. The number of rounds, $T_n$, is dictated by a cost budget, to be introduced below. The corresponding observation is given as
\begin{align}
    y^{(m)}_{n,t} = f_n^{(m)}(\mathbf{x}_{n,t})+\epsilon_{n,t}, \label{eq: noisy obs}
\end{align}
where the observation noise variables $\epsilon_{n,t}\sim\mathcal{N}(0,\sigma^2)$ are independent. Each pair $(\mathbf{x}_{n,t},m_{n,t})$ is chosen by the optimizer based on the past observations
\begin{align}
    \mathcal{D}_{n,t}=\Big\{(\mathbf{x}_{n,1},m_{n,1},y_{n,1}^{(m_1)}),...,(\mathbf{x}_{n,t},m_{n,t},y_{n,t}^{(m_t)})\Big\} \label{eq: current dataset}
\end{align}
for the current task $n$, as well as based on the dataset
\begin{align}
    \mathcal{D}_{n-1}=\bigcup_{n'=1}^{n-1} \mathcal{D}_{n',T_{n'}}
\end{align}
collected for all the previous tasks $n'=1,...,n-1$. In practice, as we will see, data set $\mathcal{D}_{n-1}$ need not be explicitly stored. Rather, information in $\mathcal{D}_{n-1}$ that is useful for future tasks is summarized into a distribution over the shared parameter vector $\boldsymbol{\theta}$. 

The number of rounds $T_n$ is determined by the \emph{cost constraint}
\begin{align}
    \sum_{t=1}^{T_n}\lambda^{(m_t)}\leq\Lambda \label{eq: budget}
\end{align}
for each task $n$, where $\Lambda$ is a pre-determined total query cost budget for each optimization task. Accordingly, the number of rounds $T_n$ is the maximum integer such that constraint \eqref{eq: budget} is satisfied.

\subsection{Gaussian Process}\label{ssec: sogp}
The proposed approach builds on multi-fidelity GPs. To explain, we begin in this subsection with a brief review of \emph{conventional GP}, which corresponds to the special case $M=1$ of a \emph{single fidelity level}. With $M=1$, the optimizer maintains a single surrogate objective function for the true objective $f^{(1)}_n(\mathbf{x})=f_n(\mathbf{x})$ of each task $n$. In BO, this is done by assigning a zero-mean GP distribution $p(f_n|\boldsymbol{\theta})$ to function $f_n(\mathbf{x})$ that is characterized by a \emph{kernel function} $k_{\boldsymbol{\theta}}(\mathbf{x},\mathbf{x}')$, as denoted by
\begin{align}
    f_n\sim \mathcal{GP}(0,k_{\boldsymbol{\theta}}(\mathbf{x}, \mathbf{x}')).\label{eq: gp prior}
\end{align}
The notation $k_{\boldsymbol{\theta}}(\mathbf{x},\mathbf{x}')$ makes it clear that the kernel function, measuring the correlation of function values $f_n(\mathbf{x})$ and $f_n(\mathbf{x}')$, depends on the common parameters $\boldsymbol{\theta}$ in \eqref{eq: theta prior}.

By definition of GP, the collection of objective values $\mathbf{f}_{n,t}=[f_n(\mathbf{x}_{n,1}),...,f_n(\mathbf{x}_{n,t})]$ from any set of inputs $\mathbf{X}_{n,t}=[\mathbf{x}_{n,1},...,\mathbf{x}_{n,t}]$ follows a multivariate Gaussian distribution $\mathcal{N}(\mathbf{0},\mathbf{K}_{\boldsymbol{\theta}}(\mathbf{X}_{n,t}))$, with $t\times 1$ zero mean vector $\mathbf{0}$, and $t\times t$ covariance matrix $\mathbf{K}_{\boldsymbol{\theta}}(\mathbf{X}_{n,t})$ given by
\begin{align}
    \mathbf{K}_{\boldsymbol{\theta}_n}(\mathbf{X}_{n,t})=\begin{bmatrix}
k_{\boldsymbol{\theta}_n}(\mathbf{x}_{n,1},\mathbf{x}_{n,1}) & ... & k_{\boldsymbol{\theta}_n}(\mathbf{x}_{n,1},\mathbf{x}_{n,t})\\
\vdots & \ddots & \vdots \\ k_{\boldsymbol{\theta}_n}(\mathbf{x}_{n,t},\mathbf{x}_{n,1}) & ... & k_{\boldsymbol{\theta}_n}(\mathbf{x}_{n,t},\mathbf{x}_{n,t})
\end{bmatrix}. \label{eq: covar mat}
\end{align}
A typical parametric kernel function is given by \cite{rothfuss2021pacoh}
\begin{align}
    k_{\boldsymbol{\theta}}(\mathbf{x},\mathbf{x}')=\exp{(-||\psi_{\boldsymbol{\theta}}(\mathbf{x})-\psi_{\boldsymbol{\theta}}(\mathbf{x}')||_2^2}),\label{eq: sogp para kernel}
\end{align}
where $\psi_{\boldsymbol{\theta}}(\cdot)$ is a neural network with parameters $\boldsymbol{\theta}$.

Given any observation history $\mathcal{D}_{n,t}$ for task $n$ in \eqref{eq: current dataset}, and given a parameter vector $\boldsymbol{\theta}$, the posterior distribution of objective value $f_n(\mathbf{x})$ at any input $\mathbf{x}$ is the Gaussian distribution \cite{Rasmussen2004}
\begin{equation}p_{\boldsymbol{\theta}}(f_n(\mathbf{x})|\mathcal{D}_{n,t})= \mathcal{N}(\mu_{\boldsymbol{\theta}}(\mathbf{x}|\mathcal{D}_{n,t}), \sigma_{\boldsymbol{\theta}}^2(\mathbf{x}|\mathcal{D}_{n,t})),\label{eq: sogp posterior}\end{equation} 
where
\begin{subequations}
    \begin{equation}
    \mu_{\boldsymbol{\theta}}(\mathbf{x}|\mathcal{D}_{n,t})=\mathbf{k}_{\boldsymbol{\theta}}(\mathbf{x})^{\sf T}(\tilde{\mathbf{K}}_{\boldsymbol{\theta}}(\mathbf{X}_{n,t}))^{-1}\mathbf{y}_{n,t},\label{eq: sogp mean}\end{equation}
    \begin{equation} \text{and}\quad\sigma_{\boldsymbol{\theta}}^2(\mathbf{x}|\mathcal{D}_{n,t})=k_{\boldsymbol{\theta}}(\mathbf{x},\mathbf{x})-\mathbf{k}_{\boldsymbol{\theta}}(\mathbf{x})^{\sf T}(\tilde{\mathbf{K}}_{\boldsymbol{\theta}}(\mathbf{X}_{n,t}))^{-1}\mathbf{k}_{\boldsymbol{\theta}}(\mathbf{x}),\label{eq: sogp var}\end{equation}
\end{subequations}
with the $t\times t$ Gramian matrix $\tilde{\mathbf{K}}_{\boldsymbol{\theta}}(\mathbf{X}_{n,t})=\mathbf{K}_{\boldsymbol{\theta}}(\mathbf{X}_{n,t})+\sigma^2\mathbf{I}_{t}$; the $t \times 1$ cross-variance vector $\mathbf{k}_{\boldsymbol{\theta}}(\mathbf{x})=[k_{\boldsymbol{\theta}}(\mathbf{x},\mathbf{x}_{n,1}),\hdots,k_{\boldsymbol{\theta}}(\mathbf{x},\mathbf{x}_{n,t})]^{\sf T}$; and the $t\times 1$ observations vector $\mathbf{y}_{n,t}=[y_{n,1},...,y_{n,t}]^{\sf T}$.

\subsection{Multi-fidelity Gaussian Process}\label{ssec: mfgp}
\emph{Multi-fidelity GP} (MFGP) provides a surrogate model for the objective functions $(f_n^{(1)}(\mathbf{x}),...,f_n^{(M)}(\mathbf{x}))$ across all $M$ fidelity levels \cite{bonilla2007multi,kennedy2000predicting}. This is done by defining a kernel function of the form $k_{\boldsymbol{\theta}}\big((\mathbf{x},m),(\mathbf{x}',m')\big)$ that captures the correlations between the function values $f_n^{(m)}(\mathbf{x})$ and $f_n^{(m')}(\mathbf{x}')$ for any two inputs $\mathbf{x}$ and $\mathbf{x}'$, and for any two fidelity levels $m$ and $m'$. Examples of such kernels include the co-kriging model in \cite{kennedy2000predicting} and the \emph{intrinsic coregionalization model} (ICM) kernel \cite{bonilla2007multi,alvarez2011computationally}, which is expressed as
\begin{align}
    k_{\boldsymbol{\theta}}\big((\mathbf{x},m),(\mathbf{x}',m')\big)=k_{\boldsymbol{\theta}}(\mathbf{x},\mathbf{x}')\cdot \kappa_{\boldsymbol{\theta}}(m,m'),\label{eq: icm kernel}
\end{align}
where the input-space kernel $k_{\boldsymbol{\theta}}(\mathbf{x},\mathbf{x}')$ is defined as in the previous subsection; while the fidelity space kernel $\kappa_{\boldsymbol{\theta}}(m,m')$ is often instantiated as a \emph{radial basis function} (RBF) kernel
\begin{align}
    \kappa(m,m')=\exp(-\gamma||m-m'||^2),\label{eq: rbf fidelity kernel}
\end{align}
where the bandwidth parameter $\gamma$ is included in the hyperparameters $\boldsymbol{\theta}$. The MFGP prior for functions $f_n^{(1)}(\mathbf{x}),...,f_n^{(M)}(\mathbf{x})$ is denoted as 
\begin{align}
    \Big(f_n^{(1)}(\mathbf{x}),...,f_n^{(M)}(\mathbf{x})\Big)\sim\mathcal{GP}(0,k_{\boldsymbol{\theta}}((\mathbf{x},m),(\mathbf{x}',m'))). \label{eq: mfgp sample}
\end{align}

Let $\mathbf{y}_{n,t}=[y^{(m_1)}_{n,1},...,y^{(m_{t})}_{n,t}]^{\sf T}$ denote the $t\times 1$ observations vector, and $\mathbf{m}_{n,t}=[m_{n,1},...,m_{n,t}]^{\sf T}$ be the $t\times 1$ vector of queried fidelity levels. Using the $t\times t$ kernel matrix $\mathbf{K}_{\boldsymbol{\theta}}(\mathbf{X}_{n,t},\mathbf{m}_{n,t})$ in which the $(i,j)$-th element $k_{\boldsymbol{\theta}}((\mathbf{x}_{n,i},m_{n,i}),(\mathbf{x}_{n,j},m_{n,j}))$ is defined as in \eqref{eq: icm kernel}, the MFGP posterior mean and variance at any input $\mathbf{x}$ with fidelity $m$ given the observation history $\mathcal{D}_{n,t}$ can be expressed as \cite{takeno2020multi}
\begin{subequations}
    \begin{equation}
        \mu^{(m)}_{\boldsymbol{\theta}}(\mathbf{x}|\mathcal{D}_{n,t})=\mathbf{k}_{\boldsymbol{\theta}}(\mathbf{x},m)^{\sf T}(\tilde{\mathbf{K}}_{\boldsymbol{\theta}}(\mathbf{X}_{n,t},\mathbf{m}_{n,t}))^{-1}\mathbf{y}_{n,t},\label{eq: mtgp mean}
    \end{equation}
    \begin{align}
        [\sigma_{\boldsymbol{\theta}}^{(m)}(\mathbf{x}|\mathcal{D}_{n,t})]^2=&k_{\boldsymbol{\theta}}((\mathbf{x},m),(\mathbf{x},m))\nonumber\\&\hspace{-0.06cm}-\mathbf{k}_{\boldsymbol{\theta}}(\mathbf{x},m)^{\sf T}(\tilde{\mathbf{K}}_{\boldsymbol{\theta}}(\mathbf{X}_{n,t},\mathbf{m}_{n,t}))^{-1}\mathbf{k}_{\boldsymbol{\theta}}(\mathbf{x},m),\label{eq: mtgp variance}
    \end{align}\label{eq: mfgp posterior}
\end{subequations}
where $\tilde{\mathbf{K}}_{\boldsymbol{\theta}}(\mathbf{X}_{n,t},\mathbf{m}_{n,t})=\mathbf{K}_{\boldsymbol{\theta}}(\mathbf{X}_{n,t},\mathbf{m}_{n,t})+\sigma^2\mathbf{I}_{t}$ is the $t\times t$ Gramian matrix. 

Based on \eqref{eq: mfgp posterior}, an estimated value of the objective $f^{(m)}_n(\mathbf{x})$ can be obtained as the mean $\mu^{(m)}_{\boldsymbol{\theta}}(\mathbf{x}|\mathcal{D}_{n,t})$ in \eqref{eq: mtgp mean}, and the corresponding uncertainty of the estimate can be quantified by the variance $[\sigma_{\boldsymbol{\theta}}^{(m)}(\mathbf{x}|\mathcal{D}_{n,t})]^2$ in \eqref{eq: mtgp variance}.

\section{Single-Task Multi-fidelity Bayesian Optimization}\label{sec: single mfbo}
In this section, we review a baseline implementation of MFBO based on MES \cite{wang2017max,moss2021gibbon}, which applies separately to each task $n$, without attempting to transfer knowledge across tasks. 

\subsection{Multi-Fidelity Max-Value Entropy Search}\label{ssec: mfmes detail}
For brevity of notation, we henceforth omit the dependence on the observation history $\mathcal{D}_{n,t}$ of the MFGP posterior mean $\mu^{(m)}_{\boldsymbol{\theta}}(\mathbf{x}|\mathcal{D}_{n,t})$ and variance $[\sigma_{\boldsymbol{\theta}}^{(m)}(\mathbf{x}|\mathcal{D}_{n,t})]^2$ in \eqref{eq: mfgp posterior}, writing $\mu^{(m)}_{\boldsymbol{\theta}}(\mathbf{x})$ and $[\sigma^{(m)}_{\boldsymbol{\theta}}(\mathbf{x})]^2$, respectively. Throughout this subsection, the parameter vector $\boldsymbol{\theta}$ is fixed, and the selection of $\boldsymbol{\theta}$ is discussed in the next subsection. In general \emph{multi-fidelity max-value entropy search} (MF-MES) \cite{moss2021gibbon}, at each time $t$, the next input $\mathbf{x}_{n,t+1}$ is selected, together with the fidelity level $m_{n,t+1}$, so as to maximize the ratio between the \emph{informativeness} of the resulting observation $y_n^{(m)}$ in \eqref{eq: noisy obs} and the cost $\lambda^{(m)}$. 

Informativeness is measured by the \emph{mutual information} between the optimal value $f_n(\mathbf{x}_n^*)=f_n^*$ of the objective and the observation $y_{n}^{(m)}$ corresponding to input $\mathbf{x}_{n,t+1}$ at fidelity level $m$. Accordingly, the next pair $(\mathbf{x}_{n,t+1},m_{n,t+1})$ is obtained by maximizing the information gain per cost unit as
\begin{align}
    (\mathbf{x}_{n,t+1},m_{n,t+1})=\arg\max\limits_{\substack{\mathbf{x}\in\mathcal{X}\\m\in\mathcal{M}}}\frac{I(f_n^*;y_{n}^{(m)}|\mathbf{x},\boldsymbol{\theta},\mathcal{D}_{n,t})}{\lambda^{(m)}}.\label{eq: argmax mfmes}
\end{align}
In \eqref{eq: argmax mfmes}, the mutual information is evaluated with respect to the joint distribution
\begin{align}
    &p(f_n^*,f_n^{(m)}(\mathbf{x}),y_n^{(m)}|\mathbf{x},\boldsymbol{\theta},\mathcal{D}_{n,t})\nonumber\\&\hspace{-0.1cm}=p(f_n^*,f_n^{(m)}(\mathbf{x})|\mathbf{x},\boldsymbol{\theta},\mathcal{D}_{n,t})p(y_n^{(m)}|f_n^{(m)}(\mathbf{x})),\label{eq: joint dist}
\end{align}
where $p(f_n^*,f_n^{(m)}(\mathbf{x})|\mathbf{x},\boldsymbol{\theta},\mathcal{D}_{n,t})$ follows the posterior GP in \eqref{eq: mfgp posterior}, and $p(y_n^{(m)}|f_n^{(m)}(\mathbf{x}))$ is defined by the observation model \eqref{eq: noisy obs}. Note that, to evaluate \eqref{eq: argmax mfmes}, the true, unobserved, function value $f_n^{(m)}(\mathbf{x})$ must be marginalized over.

Let us write as $H(\mathbf{y}|\mathbf{x})$ for the differential entropy of a variable $\mathbf{y}$ given a variable $\mathbf{x}$. By definition, the mutual information in \eqref{eq: argmax mfmes} can be expressed as the difference of differential entropies \cite{tm2006elements}
\begin{align}
    &I(f_n^*;y_{n}^{(m)}|\mathbf{x},\boldsymbol{\theta},\mathcal{D}_{n,t})\nonumber\\&\hspace{-0.1cm}=H(y_{n}^{(m)}|\mathbf{x},\boldsymbol{\theta},\mathcal{D}_{n,t})\nonumber\\& \quad-\mathbbm{E}_{p(f_n^*|\mathbf{x},\boldsymbol{\theta},\mathcal{D}_{n,t})}[H(y_{n}^{(m)}|f_n^*,\mathbf{x},\boldsymbol{\theta},\mathcal{D}_{n,t})]   \nonumber\\ &\hspace{-0.1cm}=\log(\sqrt{2\pi e}\sigma_{\boldsymbol{\theta}}^{(m)}(\mathbf{x}))\nonumber\\&\quad-\mathbbm{E}_{p(f_n^*|\boldsymbol{\theta},\mathcal{D}_{n,t})}[H(y_{n}^{(m)}|f_n^{(m)}(\mathbf{x})\leq f_n^*,\mathbf{x},\boldsymbol{\theta},\mathcal{D}_{n,t})],\label{eq: mfmes closed}
\end{align}
where the variance $[\sigma_{\boldsymbol{\theta}}^{(m)}(\mathbf{x})]^2$ is as in \eqref{eq: mtgp variance}, and \eqref{eq: mfmes closed} relies on the assumption that the $m$-th surrogate function $f_n^{(m)}(\mathbf{x})$ cannot attain a value larger than the maximum $f_n^*$ of the true objective function $f_n(\mathbf{x})$. Alternatively, one can remove this assumption by adopting a more complex approximation illustrated as in \cite{takeno2020multi}. In MF-MES, the second term in \eqref{eq: mfmes closed} is approximated as \cite{moss2021gibbon}
\begin{align}
    &H(y_{n}^{(m)}|f_n^{(m)}(\mathbf{x})\leq f_n^*,\mathbf{x},\boldsymbol{\theta},\mathcal{D}_{n,t})\nonumber \\&\hspace{-0.1cm}\approx\log\Bigg(\sqrt{2\pi e}\sigma_{\boldsymbol{\theta}}^{(m)}(\mathbf{x})\bigg(1-\frac{\phi(\gamma_{\boldsymbol{\theta}}^{(m)}(\mathbf{x},f_n^*))}{\Phi(\gamma_{\boldsymbol{\theta}}^{(m)}(\mathbf{x},f_n^*))}\nonumber\\&\quad\bigg[\gamma_{\boldsymbol{\theta}}^{(m)}(\mathbf{x},f_n^*)+\frac{\phi(\gamma_{\boldsymbol{\theta}}^{(m)}(\mathbf{x},f_n^*))}{\Phi(\gamma_{\boldsymbol{\theta}}^{(m)}(\mathbf{x},f_n^*))}\bigg]\bigg)\Bigg)\nonumber\\&\hspace{-0.1cm}=H_{\boldsymbol{\theta}}^{\text{MF-MES}}(\mathbf{x},m,f^*_n)\label{eq: esg closed} \\ &\text{with} \quad \gamma_{\boldsymbol{\theta}}^{(m)}(\mathbf{x},f_n^*)=\frac{f_n^*-\mu_{\boldsymbol{\theta}}^{(m)}(\mathbf{x})}{\sigma_{\boldsymbol{\theta}}^{(m)}(\mathbf{x})},\label{eq: gamma func}
\end{align}
where $\phi(\cdot)$ and $\Phi(\cdot)$ are the probability density function and cumulative density function of a standard Gaussian distribution, respectively. Intuitively, function $H_{\boldsymbol{\theta}}^{\text{MF-MES}}(\mathbf{x},m,f^*_n)$ in \eqref{eq: gamma func} captures the uncertainty on the observation $y_n^{(m)}$ that would be produced by querying the objective at input $\mathbf{x}$ and fidelity level $m$ when the optimal value $f_n^*$ is known. This uncertainty should be subtracted, as per \eqref{eq: mfmes closed}, from the overall uncertainty $H(y_n^{(m)}|\mathbf{x},\boldsymbol{\theta},\mathcal{D}_{n,t})$ in order to assess the extent to which the observation $y_n^{(m)}$ provides information about the optimal value $f_n^*$.

Using \eqref{eq: esg closed} in \eqref{eq: mfmes closed} and replacing the expectation in \eqref{eq: mfmes closed} with an empirical average, MF-MES selects the next query as
\begin{align}
    (\mathbf{x}_{n,t+1},m_{n,t+1})=\arg\max\limits_{\substack{\mathbf{x}\in\mathcal{X}\\m\in\mathcal{M}}}\alpha_{\boldsymbol{\theta}}^{\text{MF-MES}}(\mathbf{x},m),\label{eq: mfmes selection}
\end{align}
where we have defined the MF-MES acquisition function
\begin{align}
    \alpha_{\boldsymbol{\theta}}^{\text{MF-MES}}(\mathbf{x},m)=&\frac{1}{\lambda^{(m)}}\log(\sqrt{2\pi e}\sigma_{\boldsymbol{\theta}}^{(m)}(\mathbf{x}))\nonumber\\&-\frac{1}{|\mathcal{F}|\lambda^{(m)}}\sum_{f_n^*\in\mathcal{F}}H_{\boldsymbol{\theta}}^{\text{MF-MES}}(\mathbf{x},m,f_n^*),\label{eq: af closed}
\end{align}
where the set $\mathcal{F}=\{f_{n,s}^*\}_{s=1}^S$ collects 
$S$ samples drawn from distribution $p_{\boldsymbol{\theta}}(f_n^*|\mathbf{x},\mathcal{D}_{n,t})$, which can be obtained via Gumbel sampling (see \cite[Section 3.1]{wang2017max} for details) and function $H_{\boldsymbol{\theta}}^{\text{MF-MES}}(\mathbf{x},m,f_n^*)$ is defined in \eqref{eq: esg closed}.

The overall procedure of MF-MES is summarized in Algorithm \ref{table: mfmes}.

\begin{algorithm}[t!]
\caption{Multi-Fidelity Max-Value Entropy Search (MF-MES) \cite{moss2021gibbon}}\label{table: mfmes}
\SetKwInOut{Input}{Input}
\Input{Vector $\boldsymbol{\theta}$, simulation costs $\{\lambda^{(m)}\}_{m=1}^M$, query budget $\Lambda$}
\SetKwInOut{Output}{Output}
\Output{Optimized solution $\mathbf{x}^{\text{opt}}$}\
Initialize iteration $t=0$, observation dataset $\mathcal{D}_{n,t}=\emptyset$, and $\Lambda^0=\Lambda$\\
\While{\emph{$\Lambda^t>0$}}{
Sample max-value set $\mathcal{F}$ from distribution $p_{\boldsymbol{\theta}}(f_n^*|\mathbf{x},\mathcal{D}_{n,t})$\\
Obtain the next decision pair $(\mathbf{x}_{n,t+1},m_{n,t+1})$ via \eqref{eq: mfmes selection}\\
Observe $y_{n,t+1}^{(m)}$ in \eqref{eq: noisy obs} and update observation history $\mathcal{D}_{n,t+1}=\mathcal{D}_{n,t}\cup(\mathbf{x}_{n,t+1},m_{n,t+1},y^{(m_{n,t+1})}_{n,t+1})$\\
Update the MFGP posterior as in \eqref{eq: mfgp posterior}\\
Calculate remaining budget $\Lambda^{(t+1)}=\Lambda^{(t)}-\lambda^{(m_{n,t+1})}$\\
Set iteration $T_n=t$ and $t=t+1$
}
Return $\mathbf{x}^{\text{opt}}=\arg\max\limits_{t=1,..,T_n}f^{(M)}_n(\mathbf{x}_{n,t})$\\
\end{algorithm}

\subsection{Optimizing the Parameter Vector $\boldsymbol{\theta}$}\label{ssec: opt theta}
In Sec. \ref{ssec: mfmes detail}, we have treated the parameter vector $\boldsymbol{\theta}$ as fixed. In practice, given the data set $\mathcal{D}_{n,t}$ collected up to round $t$ for the current task $n$, it is possible to update the parameter vector $\boldsymbol{\theta}$ to fit the available observations \cite{Rasmussen2004}. This is typically done by maximizing the marginal likelihood of parameters $\boldsymbol{\theta}$ given data $\mathcal{D}_{n,t}$.

Under the posterior distribution defined by \eqref{eq: mtgp mean} and \eqref{eq: mtgp variance}, the \emph{negative marginal log-likelihood} of the parameter vector $\boldsymbol{\theta}$ is given by
\begin{align}
    \ell(\boldsymbol{\theta}|\mathcal{D}_{n,t})&=-\log\big( p(\mathbf{y}_{n,t}|\boldsymbol{\theta})\big)\nonumber\\&=-\frac{1}{2}\Bigg(t\log2\pi-\log\Big|\tilde{\mathbf{K}}_{\boldsymbol{\theta}}(\mathbf{X}_{n,t},\mathbf{m}_{n,t})\Big|\nonumber\\&\quad\,-\mathbf{y}_{n,t}^{\sf T}\Big(\tilde{\mathbf{K}}_{\boldsymbol{\theta}}(\mathbf{X}_{n,t},\mathbf{m}_{n,t})\Big)^{-1}\mathbf{y}_{n,t}\Bigg).\label{eq: mtgp mml}
\end{align}
where $p(\mathbf{y}_{n,t}|\boldsymbol{\theta})$ represents the probability density function of observation in \eqref{eq: noisy obs}. The negative marginal log-likelihood \eqref{eq: mtgp mml} can be interpreted as a loss function associated with parameters $\boldsymbol{\theta}$ based on the observations in data set $\mathcal{D}_{n,t}$. Accordingly, using the prior distribution \eqref{eq: theta prior}, a \emph{maximum a posterior} (MAP) solution for the parameter vector $\boldsymbol{\theta}$ is obtained by addressing the problem
\begin{align}
    \boldsymbol{\theta}_{n,t}^{\text{MAP}}=\arg\min\limits_{\boldsymbol{\theta}\in\boldsymbol{\Theta}}\{\ell(\boldsymbol{\theta}|\mathcal{D}_{n,t})-\log \big(p(\boldsymbol{\theta})\big)\},\label{eq: single task theta}
\end{align}
where $p(\boldsymbol{\theta})$ is the prior distribution in \eqref{eq: theta prior}, and the term $\log \big(p(\boldsymbol{\theta})\big)$ plays the role of a regularizer. To reduce computational complexity, the optimization problem \eqref{eq: single task theta} may be addressed periodically with respect to the round index $t$ \cite{damianou2013deep,Maggi2021bogp}.

\section{Sequential Multi-fidelity Bayesian Optimization With Transferable Max-Value Entropy Search}\label{sec: k-mfbo}
As reviewed in the previous section, MF-MES treats each task $n$ separately \cite{takeno2020multi,moss2021gibbon}. However, since by \eqref{eq: sample f_n} and \eqref{eq: theta prior}, the successive objective functions $f_1,...,f_n$ are generally correlated, knowledge extracted from one task can be transferred to future tasks through the common parameters 
$\boldsymbol{\theta}$ \cite{sloman2023fundamental}. In this section, we introduce a novel information-theoretic acquisition function that generalizes the MF-MES acquisition function in \eqref{eq: argmax mfmes} to account for the information acquired about parameters $\boldsymbol{\theta}$ for future tasks. The proposed approach, referred to as \emph{multi-fidelity transferable max-value entropy search} (MFT-MES), hinges on a Bayesian formulation for the problem of sequentially estimating parameter vector $\boldsymbol{\theta}$ as more tasks are observed. In this section, we first describe the proposed acquisition function, and then describe an efficient implementation of the Bayesian estimation of parameter vector $\boldsymbol{\theta}$ based on \emph{Stein variational gradient descent} (SVGD) \cite{liu2016stein}.

\subsection{Multi-Fidelity Transferable Max-Value Entropy Search}\label{ssec: mf-tmes}
MFT-MES introduces a term in the MF-MES acquisition function \eqref{eq: argmax mfmes} that promotes the selection of inputs $\mathbf{x}$ and fidelity levels $m$ that maximize the information brought by the corresponding observation $y_{n,t}^{(m)}$ about the parameters $\boldsymbol{\theta}$. The rationale behind this modification is that collecting information about the shared parameters $\boldsymbol{\theta}$ can potentially improve the optimization process for future tasks $n'>n$.

Accordingly, MFT-MES adopts an acquisition function that measures the mutual information between the observation $y_n^{(m)}$ and, not only the optimal value $f_n^*$ for the current task $n$ in MF-MES, but also the common parameters $\boldsymbol{\theta}$. Note that, unlike MF-MES, this approach views the parameter vector $\boldsymbol{\theta}$ as a random quantity that is jointly distributed with the objective and with the observations. Normalizing by the cost $\lambda^{(m)}$ as in \eqref{eq: argmax mfmes}, MFT-MES selects the next query $(\mathbf{x}_{n,t+1},m_{n,t+1})$ with the aim of addressing the problem
\begin{align}
    (\mathbf{x}_{n,t+1},m_{n,t+1})=\arg\max\limits_{\substack{\mathbf{x}\in\mathcal{X}\\m\in\mathcal{M}}}\frac{I(f_n^*,\boldsymbol{\theta};y_{n}^{(m)}|\mathbf{x},\mathcal{D}_{n-1},\mathcal{D}_{n,t})}{\lambda^{(m)}},\label{eq: smfmes acq}
\end{align}
which is evaluated with respect to the joint distribution
\begin{align}
    &p(f_n^*,f_n^{(m)}(\mathbf{x}),\boldsymbol{\theta},y_n^{(m)}|\mathbf{x},\mathcal{D}_{n-1},\mathcal{D}_{n,t})\nonumber\\&\hspace{-0.1cm}=p(\boldsymbol{\theta}|\mathcal{D}_{n-1},\mathcal{D}_{n,t})p(f_n^*,f_n^{(m)}(\mathbf{x})|\mathbf{x},\boldsymbol{\theta},\mathcal{D}_{n,t})p(y_n^{(m)}|f_n^{(m)}(\mathbf{x})),\label{eq: mft joint dist}
\end{align}
where the distribution $p(\boldsymbol{\theta}|\mathcal{D}_{n-1},\mathcal{D}_{n,t})$ is the posterior distribution on the shared model parameters $\boldsymbol{\theta}$. This posterior distribution represents the knowledge extracted from all the observations available at round $t$ of task $n$  that can be transferred to future tasks. It is obtained as
\begin{align}
    p(\boldsymbol{\theta}|\mathcal{D}_{n-1},\mathcal{D}_{n,t})=\frac{p(\boldsymbol{\theta},\mathcal{D}_{n-1},\mathcal{D}_{n,t})}{p(\mathcal{D}_{n-1},\mathcal{D}_{n,t})}\label{eq: theta posterior}
\end{align}
with the joint distribution
\begin{align}
    &p(\boldsymbol{\theta},\mathcal{D}_{n-1},\mathcal{D}_{n,t})\nonumber\\&=p(\boldsymbol{\theta})\cdot\prod_{i=1}^{n-1}p(\mathcal{D}_i|\boldsymbol{\theta})p(\mathcal{D}_{n,t}|\boldsymbol{\theta})\nonumber\\&=p(\boldsymbol{\theta})\cdot\prod_{i=1}^{n-1}p(\mathbf{y}_i|\mathbf{x},\boldsymbol{\theta})p(\mathbf{x})\cdot p(\mathbf{y}_{n,t}|\mathbf{x},\boldsymbol{\theta})p(\mathbf{x}),
\end{align}
and $p(\mathcal{D}_{n-1},\mathcal{D}_{n,t})=\int p(\boldsymbol{\theta})p(\mathcal{D}_{n-1},\mathcal{D}_{n,t}|\boldsymbol{\theta})d\boldsymbol{\theta}$.

Using the chain rule for mutual information \cite{tm2006elements}, the acquisition function \eqref{eq: smfmes acq} is expressed as
\begin{align}
    &\alpha^{\text{MFT-MES}}(\mathbf{x},m)\nonumber\\&\hspace{-0.1cm}=\frac{1}{\lambda^{(m)}}\Big[\mathbbm{E}_{p(\boldsymbol{\theta}|\mathcal{D}_{n-1},\mathcal{D}_{n,t})}\big[I(f_n^*;y_{n}^{(m)}|\mathbf{x},\boldsymbol{\theta},\mathcal{D}_{n,t})\big]\nonumber\\&\quad+I(\boldsymbol{\theta};y_{n}^{(m)}|\mathbf{x},\mathcal{D}_{n-1},\mathcal{D}_{n,t})\Big].\label{eq: meta acq func}
\end{align}
In \eqref{eq: meta acq func}, the first term is the expected information gain on the global optimum $f_n^*$ that is also included in the MF-MES acquisition function in \eqref{eq: argmax mfmes}, while the second term in \eqref{eq: meta acq func} quantifies transferable knowledge via the information gain about the shared parameters $\boldsymbol{\theta}$.

The second term in \eqref{eq: meta acq func} can be written more explicitly as
\begin{align}
    &I(\boldsymbol{\theta};y_n^{(m)}|\mathbf{x},\mathcal{D}_{n-1},\mathcal{D}_{n,t})\nonumber\\ &\hspace{-0.1cm}=H(y_{n}^{(m)}|\mathbf{x},\mathcal{D}_{n-1},\mathcal{D}_{n,t})\nonumber\\&\quad-\mathbbm{E}_{p(\boldsymbol{\theta}|\mathcal{D}_{n-1},\mathcal{D}_{n,t})}\big[H(y^{(m)}_{n}|\mathbf{x},\boldsymbol{\theta},\mathcal{D}_{n,t})\big]\nonumber\\ &\hspace{-0.1cm}= 
    H\bigg(\mathbb{E}_{p(\boldsymbol{\theta}|\mathcal{D}_{n-1},\mathcal{D}_{n,t})}\Big[ p(y_{n}^{(m)}|\mathbf{x}, \boldsymbol{\theta}, \mathcal{D}_{n,t}) \Big]\bigg)\nonumber\\&\quad-\frac{1}{2}\mathbb{E}_{p(\boldsymbol{\theta}|\mathcal{D}_{n-1},\mathcal{D}_{n,t})}\Big[\log\Big(2\pi e\big[\sigma_{\boldsymbol{\theta}}^{(m)}(\mathbf{x})\big]^2\Big)\Big],\label{eq: transferable knowledge}
\end{align}
where the first term in \eqref{eq: transferable knowledge} is the differential entropy of a mixture of Gaussians. In fact, each distribution $p(y_n^{(m)}|\mathbf{x},\boldsymbol{\theta},\mathcal{D}_{n,t})$ corresponds to the GP posterior \eqref{eq: mfgp posterior} with mean $\mu^{(m)}_{\boldsymbol{\theta}}(\mathbf{x})$ and variance $[\sigma_{\boldsymbol{\theta}}^{(m)}(\mathbf{x})]^2$.

Evaluating the first term in \eqref{eq: transferable knowledge} is problematic \cite{tm2006elements}, and here we resort to the upper bound obtained via the principle of maximum entropy \cite{weiss2006course,moshksar2016arbitrarily,nielsen2016guaranteed}
\begin{align}
&H\bigg(\mathbb{E}_{p(\boldsymbol{\theta}|\mathcal{D}_{n-1},\mathcal{D}_{n,t})}\Big[ \mathcal{N}(y_n^{(m)}|\mu^{(m)}_{\boldsymbol{\theta}}(\mathbf{x}),\sigma_{\boldsymbol{\theta}}^{(m)2}(\mathbf{x})) \Big]\bigg) \nonumber\\&\leq \frac{1}{2}\log\Big(2\pi e \text{Var}(y_{n}^{(m)})\Big), \label{eq: entropy bound mix gp}
\end{align}
where $\text{Var}(y_{n}^{(m)})$ represents the variance of random variable $y^{(m)}_n$ following the mixture of Gaussian distribution $\mathbb{E}_{p(\boldsymbol{\theta}|\mathcal{D}_{n-1},\mathcal{D}_{n,t})}[ \mathcal{N}(y_n^{(m)}|\mu^{(m)}_{\boldsymbol{\theta}}(\mathbf{x}),\sigma_{\boldsymbol{\theta}}^{(m)2}(\mathbf{x})) ]$. The tightness of upper bound \eqref{eq: entropy bound mix gp} depends on the diversity of the Gaussian components with respect to the distribution $p(\boldsymbol{\theta}|\mathcal{D}_{n-1}, \mathcal{D}_{n,t})$, becoming more accurate when the components of the mixture tend to the same Gaussian distribution \cite{moshksar2016arbitrarily}.

\begin{algorithm}[t!]
\caption{Multi-Fidelity Transferable Max-Value Entropy Search (MFT-MES)}\label{table: mftmes}
\SetKwInOut{Input}{Input}
\Input{Prior $p(\boldsymbol{\theta})$, scaling parameter $\beta$, simulation costs $\{\lambda^{(m)}\}_{m=1}^M$, query budget $\Lambda$, stepsize $\eta$, number of SVGD iterations $R$}
\SetKwInOut{Output}{Output}
\Output{Optimized solution $\mathbf{x}^{\text{opt}}$}\
Initialize $t=0$, $r=0$, observation dataset $\mathcal{D}_{n,t}=\emptyset$ and $\Lambda^0=\Lambda$\\
\If{$n=1$}{
Generate particles $\{\boldsymbol{\theta}_v\}_{v=1}^{V}$ i.i.d. from the prior $p(\boldsymbol{\theta})$ 
}
\Else{
Set particles $\{\boldsymbol{\theta}_v\}_{v=1}^{V}$ to the particles $\{\boldsymbol{\theta}_v^{(R)}\}_{v=1}^V$ produced from the previous task $n-1$
}
\While{\emph{$\Lambda^t>0$}}{
\For{$v\leq V$}{
Obtain max-value set $\mathcal{F}$ from distribution $p(f_n^*|\mathbf{x},\boldsymbol{\theta}_v,\mathcal{D}_{n,t})$ using Gumbel sampling\\}
Evaluate the transferable knowledge criterion $\Delta\alpha^{\text{MFT-MES}}(\mathbf{x},m)$ using \eqref{eq: upper bound trans knowledge}\\
Obtain the next decision pair $(\mathbf{x}_{n,t+1},m_{n,t+1})$ via \eqref{eq: final af}\\
Observe $y_{n,t+1}^{(m)}$ in 
\eqref{eq: noisy obs} and update observation history $\mathcal{D}_{n,t+1}=\mathcal{D}_{n,t}\cup(\mathbf{x}_{n,t+1},m_{n,t+1},y^{(m_{n,t+1})}_{n,t+1})$\\
Update the MFGP posterior as in \eqref{eq: mfgp posterior} \\
Calculate remaining budget $\Lambda^{(t+1)}=\Lambda^{(t)}-\lambda^{(m_{n,t+1})}$\\
Set iteration $T_n=t$ and $t=t+1$
}
Evaluate the negative marginal log-likelihood $\ell(\boldsymbol{\theta}_v|\mathcal{D}_{n,T_n})$ for all particles $v=1,...,V$ using \eqref{eq: mtgp mml}\\ 
\For{$r\leq R$}{
Evaluate the function $\Omega(\boldsymbol{\theta}_v^{(r)})$ as in \eqref{eq: smooth func}\\
Update each particle via the SVGD update $\boldsymbol{\theta}_v^{(r+1)} = \boldsymbol{\theta}_v^{(r)}+\eta\Omega(\boldsymbol{\theta}_v^{(r)})$
}
Return $\mathbf{x}^{\text{opt}}=\arg\max\limits_{t=1,..,T_n}f^{(M)}_n(\mathbf{x}_{n,t})$\\
\end{algorithm}

When the distribution $p(\boldsymbol{\theta}|\mathcal{D}_{n-1},\mathcal{D}_{n,t})$ is represented using $V$ particles $\{\boldsymbol{\theta}_1,...,\boldsymbol{\theta}_V\}$, the variance in \eqref{eq: entropy bound mix gp} can be approximated via the asymptotically consistent estimate
\begin{align}
\text{Var}(y_{n}^{(m)})=&\sum_{v=1}^V\frac{1}{V}\big([\sigma^{(m)}_{\boldsymbol{\theta}_v}(\mathbf{x})]^2+[\mu_{\boldsymbol{\theta}_v}^{(m)}(\mathbf{x})]^2\big)\nonumber\\&-\Bigg(\frac{1}{V}\sum_{v=1}^V\mu_{\boldsymbol{\theta}_v}^{(m)}(\mathbf{x})\Bigg)^2+\sigma^2.\label{eq: total variance}
\end{align}
Consequently, by plugging \eqref{eq: entropy bound mix gp} into \eqref{eq: transferable knowledge}, the second term in \eqref{eq: meta acq func} is replaced by the quantity
\begin{align}
    \Delta\alpha^{\text{MFT-MES}}(\mathbf{x},m)= &\frac{1}{2}\Bigg\{\log\Bigg(2\pi e\bigg(\frac{1}{V}\sum_{v=1}^V\big([\sigma^{(m)}_{\boldsymbol{\theta}_v}(\mathbf{x})]^2\nonumber\\&\hspace{-0.05cm}+[\mu_{\boldsymbol{\theta}_v}^{(m)}(\mathbf{x})]^2\big)-\Big(\frac{1}{V}\sum_{v=1}^V\mu_{\boldsymbol{\theta}_v}^{(m)}(\mathbf{x})\Big)^2\bigg)\Bigg)\nonumber \\ &\hspace{-0.05cm}-\frac{1}{V}\sum_{v=1}^V\log\Big(2\pi e\big[\sigma_{\boldsymbol{\theta}_v}^{(m)}(\mathbf{x})\big]^2\Big)\Bigg\}.\label{eq: upper bound trans knowledge}
\end{align}

Based on \eqref{eq: upper bound trans knowledge}, the proposed MFT-MES selects the next pair $(\mathbf{x}_{n,t+1},m_{n,t+1})$ for the current task $n$ by maximizing the criterion \eqref{eq: af closed} with \eqref{eq: upper bound trans knowledge}, i.e.,
\begin{align}
    (\mathbf{x}_{n,t+1},m_{n,t+1})=&\arg\max\limits_{\substack{\mathbf{x}\in\mathcal{X}\\m\in\mathcal{M}}}\mathbbm{E}_{p(\boldsymbol{\theta}|\mathcal{D}_{n-1},\mathcal{D}_{n,t})}[\alpha_{\boldsymbol{\theta}}^{\text{MF-MES}}(\mathbf{x},m)]\nonumber\\&+\beta\cdot\frac{\Delta\alpha^{\text{MFT-MES}}(\mathbf{x},m)}{\lambda^{(m)}},\label{eq: final af}
\end{align}
where the scaling parameter $\beta\geq0$ determines the relative weight assigned to the task of knowledge transfer for future tasks.

\subsection{Bayesian Learning for the Parameter Vector $\boldsymbol{\theta}$}\label{ssec: bl}
As explained in the previous section, MFT-MES models the shared parameter vector $\boldsymbol{\theta}$ as a random quantity jointly distributed with the GP posterior \eqref{eq: mfgp posterior} and observation model \eqref{eq: noisy obs}, which is expressed as in \eqref{eq: mft joint dist}. Furthermore, the evaluation of the MFT-MES acquisition function \eqref{eq: final af} requires the availability of $V$ particles that are approximately drawn from the posterior distribution $p(\boldsymbol{\theta}|\mathcal{D}_{n-1},\mathcal{D}_{n,t})$. In this subsection, we describe an efficient approach to obtain such particles via SVGD.

In SVGD \cite{liu2016stein}, the posterior distribution $p(\boldsymbol{\theta}|\mathcal{D}_{n-1},\mathcal{D}_{n,t})$ is approximated via a set of $V$ particles $\{\boldsymbol{\theta}_1, \boldsymbol{\theta}_2,...,\boldsymbol{\theta}_V\}$ that are iteratively transported to minimize the \emph{variational inference} (VI) objective denoted by the Kullback–Leibler (KL) divergence $\mathrm{KL}(q(\boldsymbol{\theta})||p(\boldsymbol{\theta}|\mathcal{D}_{n-1},\mathcal{D}_{n,t}))$ over a distribution $q(\boldsymbol{\theta})$ represented via particles $\{\boldsymbol{\theta}_1, \boldsymbol{\theta}_2,...,\boldsymbol{\theta}_V\}$. This is done via functional gradient descent in the reproducing kernel Hilbert space (RKHS). Under mild assumptions on the kernel derivatives and log-loss gradient in (39), SVGD using $V$ particles converges to the true posterior distribution at   rate $\mathcal{O}(1/\sqrt{V})$ \cite[Theorem 1]{balasubramanian2024improved}.

Specifically, the SVGD update for the $v$-th particle at each gradient descent round $r+1$ is expressed as
\begin{align}
    \boldsymbol{\theta}_v^{(r+1)} = \boldsymbol{\theta}_v^{(r)}+\eta\Omega(\boldsymbol{\theta}_v^{(r)}),\label{eq: particle update}
\end{align}
where $\eta$ is the stepsize, and the function $\Omega(\boldsymbol{\theta}_v^{(r)})$ is defined as
\begin{align}
    &\Omega(\boldsymbol{\theta}_v^{(r)})\nonumber\\&\hspace{-0.1cm}=\frac{1}{V}\sum_{v'=1}^V \Bigl[ \tilde{k}(\boldsymbol{\theta}_{v'}^{(r)},\boldsymbol{\theta}_v^{(r)})\underbrace{\nabla_{\boldsymbol{\theta}_{v'}^{(r)}} \Big(\ell(\boldsymbol{\theta}_{v'}^{(r)}|\mathcal{D}_{n,t})+\log p_n(\boldsymbol{\theta}_{v'}^{(r)})\Big)}_{\text{log-loss gradient}}\nonumber\\&+\underbrace{\nabla_{\boldsymbol{\theta}_{v'}^{(r)}} \tilde{k}(\boldsymbol{\theta}_{v'}^{(r)},\boldsymbol{\theta}_v^{(r)})}_{\text{repulsive force}}\Bigr],\label{eq: smooth func}
\end{align}
while $\tilde{k}(\cdot,\cdot)$ is a kernel function, the loss function $\ell(\boldsymbol{\theta}_{v'}^{(r)}|\mathcal{D}_{n,t})$ is as in \eqref{eq: mtgp mml} and $p_n(\boldsymbol{\theta})$ is the prior distribution at current task $n$. The first term in \eqref{eq: smooth func} drives particles $\boldsymbol{\theta}_v$ to asymptotically converge to the MAP solution \eqref{eq: single task theta}, while the second term in \eqref{eq: smooth func} is a repulsive force that maintains the diversity of particles by minimizing the similarity measure $\tilde{k}(\cdot,\cdot)$. The inclusion of the second term allows the variational distribution $q(\boldsymbol{\theta})$ represented by particles $\{\boldsymbol{\theta}_1,...,\boldsymbol{\theta}_V\}$ to capture the multi-modality of the target posterior $p(\boldsymbol{\theta}|\mathcal{D}_{n-1},\mathcal{D}_{n,t})$, providing a more accurate approximation \cite{simeone2022machine}.

The prior $p_n(\boldsymbol{\theta})$ for the current task $n$ is obtained based on the kernel density estimation (KDE) of the posterior distribution $p(\boldsymbol{\theta}|\mathcal{D}_{n-1})$ obtained by using particles $\{\boldsymbol{\theta}_1,...,\boldsymbol{\theta}_V\}$ produced at the end of the previous task $n-1$ \cite{bishop2006pattern}. The overall procedure of MFT-MES is summarized in Algorithm \ref{table: mftmes}.

\section{Theoretical Analysis}\label{sec: theorems}
In this section, we establish an upper bound on the expected regret for a simplified version of MFT-MES by building on the expected regret bound presented in \cite{wang2017max} for MES.

Throughout, we assume as in \cite{wang2017max}, that the objective functions $f_n(\mathbf{x})$ follow the GP \eqref{eq: gp prior} with a given ground-truth parameter vector $\boldsymbol{\theta}^*$. Moreover, for each task $n$, we evaluate the performance at time $t$ via average regret
\begin{align}
    \bar{r}_{n,t}(\mathbf{x}_{n,t}|\boldsymbol{\theta}^*)=\mathbbm{E}[f_n^*-f_n(\mathbf{x}_{n,t})|\mathcal{D}_{n,t-1}],\label{eq: expected regret}
\end{align}
where the expectation is taken with respect to the posterior distribution \eqref{eq: mfgp posterior} with $\boldsymbol{\theta}=\boldsymbol{\theta}^*$. The cumulative expected regret up to time $T$ for a task $n$ is accordingly defined as
\begin{align}
    \mathcal{R}_{n,T}(\mathbf{X}_{n,T}|\boldsymbol{\theta}^*)=\sum_{t=1}^T\bar{r}_{n,t}(\mathbf{x}_{n,t}|\boldsymbol{\theta}^*).\label{eq: cumulative regret}
\end{align}

\subsection{Two-Phase MFT-MES}\label{ssec: seq mes unknown}
To facilitate the analysis of the proposed MFT-MES scheme, we study a generally less efficient variant that retains the essential features of the approach. Fixing the fidelity level $M$, the simplified MFT-MES protocol under study splits the iterations for each task $n$ into two \emph{separate} phases, one devoted to per-task exploration via MES, and the other dedicated to the acquisition of information about the parameter vector $\boldsymbol{\theta}^*$. Note that MFT-MES in Algorithm \ref{table: mftmes} adopts a more flexible strategy that integrates these two phases. Appendix \ref{app: theorem 1 plot} elaborates on the impact of the assumed simplifications.

To elaborate, for each task $n$, we divide the total number of iterations $T$ into two parts via the selection of a hyperparameter $c\in [0,1]$. For the first $\lceil c\cdot T\rceil$ iterations, the next candidate solution is selected using MES as 
\begin{align} \label{eq:two_phase_MES}
    \mathbf{x}_{n,t+1} = \arg\max\limits_{\mathbf{x}\in\mathcal{X}}\mathbb{E}_{\boldsymbol{\theta}\sim p(\boldsymbol{\theta}|\mathcal{D}_{n-1})}\big[ \bar{\alpha}_{\boldsymbol{\theta}}^\text{MES}(\mathbf{x})\big]
\end{align}
for $t+1\leq c\cdot T$, where $\bar{\alpha}_{\boldsymbol{\theta}}^\text{MES}(\mathbf{x})=g(\frac{\hat{m}_{n,t}(\boldsymbol{\theta}^*)  - \mu_{\boldsymbol{\theta}^*}(\mathbf{x}) }{\sigma_{\boldsymbol{\theta}^*}(\mathbf{x})})$ is the acquisition function of MES with $\hat{m}_{n,t}(\boldsymbol{\theta}^*) = \mathbb{E}_{f_n(\mathbf{x})}[ f_n^* ]$ being the expected maximum value of the function \cite{wang2017max}. Note that, unlike the original MES scheme, the acquisition procedure \eqref{eq:two_phase_MES} applies an average over the current posterior distribution $p(\boldsymbol{\theta}|\mathcal{D}_{n-1})$ for the parameter vector $\boldsymbol{\theta}$, given that the ground-truth $\boldsymbol{\theta}^*$ is unknown.

In the second set of $T-\lceil c\cdot T\rceil$ iterations, the optimizer attempts to maximize the information gain about parameter vector $\boldsymbol{\theta}^*$ by selecting the batch of solutions $\mathbf{X}=\{ \mathbf{x}_{n,t}\}_{t=\lceil c\cdot T \rceil + 1}^{T}$ which maximize the mutual information
\begin{align} \label{eq:batch_MES}
    \{\mathbf{x}_{n,t}\}_{t= \lceil c\cdot T \rceil + 1}^{T} = \arg\max\limits_{\mathbf{X}\subset\mathcal{X}} I(\boldsymbol{\theta};\mathbf{y}|\mathbf{X}),
\end{align} 
where $\mathbf{y}$ is the  observation vector of size $T - \lceil c\cdot T \rceil$ corresponding to the candidates $\mathbf{X}$. Note that, the objective (\ref{eq:batch_MES}) simplifies \eqref{eq: transferable knowledge} in MFT-MES by (\emph{i}) selecting the $T-\lceil c\cdot T\rceil$ candidates as a batch, rather than sequentially; and (\emph{ii}) disregarding the data $\mathcal{D}_{n-1}$ collected for previous tasks.

\subsection{Technical Assumptions}\label{ssec: technical assumptions}
To describe the technical assumptions on which the analysis is based, we first put forth two definitions. These will be used to account for the gap between MES with known parameter vector $\boldsymbol{\theta}^*$ and the acquisition function \eqref{eq:two_phase_MES}, which relies on the posterior distribution $p(\boldsymbol{\theta}|\mathcal{D}_{n-1})$.

\begin{definition}[$\varepsilon$-sharpness]
      For a random parameter vector $\boldsymbol{\theta}\sim \mathcal{N}(\boldsymbol{\theta}^*, \Sigma)$, a function $h(\mathbf{x}|\boldsymbol{\theta})$ is said to be $\varepsilon$-sharp around $\boldsymbol{\theta}^*$ if the inequality
    \begin{align}
        &\Big|\Big|\arg\min_{\mathbf{x}\in\mathcal{X}} \mathbb{E}_{\boldsymbol{\theta}\sim \mathcal{N}(\boldsymbol{\theta}^*, \Sigma)}\big[ h(\mathbf{x}|\boldsymbol{\theta}) \big] - \arg\min_{\mathbf{x}\in\mathcal{X}} h(\mathbf{x}|\boldsymbol{\theta}^*) \Big|\Big|\nonumber\\& \leq \varepsilon \cdot \mathrm{Tr}(\Sigma)\label{eq: epsilon sharpness}
    \end{align}
    holds, where $\mathrm{Tr}(\cdot)$ is the trace operator. 
    \label{def:1}
\end{definition}
The $\varepsilon$-sharpness property \eqref{eq: epsilon sharpness} describes how sensitive the minimizer of the function $h(\mathbf{x}|\boldsymbol{\theta}^*)$ is with respect to the perturbations applied to the parameter vector $\boldsymbol{\theta}^*$ (see, e.g., \cite{abbas2022sharp} for related definitions).

\begin{definition}[$\varepsilon$-stability]\label{def:2} Consider the monotonically decreasing function
\begin{align} \label{eq:mono}
    g(u) = u \cdot \frac{\phi(u)}{2\Phi(u)} - \log(\Phi(u)),
\end{align}
as well as a random parameter vector $\boldsymbol{\theta}\sim\mathcal{N}(\boldsymbol{\theta}^*,\Sigma)$. The function $h(\mathbf{x}|\boldsymbol{\theta})$ is $\varepsilon$-stable around $\boldsymbol{\theta}^*$ if it satisfies the inequality 
\begin{align}
    &\Big|\Big|\arg\max\limits_{\mathbf{x}\in\mathcal{X}} \mathbb{E}_{\boldsymbol{\theta}\sim\mathcal{N}(\boldsymbol{\theta}^*,\Sigma)}\big[ g(h(\mathbf{x}|\boldsymbol{\theta}))\big]\nonumber\\&- \arg\min\limits_{\mathbf{x}\in\mathcal{X}}\mathbb{E}_{\boldsymbol{\theta}\sim\mathcal{N}(\boldsymbol{\theta}^*,\Sigma)}\big[ h(\mathbf{x}|\boldsymbol{\theta})\big] \Big|\Big|\leq\varepsilon\cdot\mathrm{Tr}(\Sigma).\label{eq: epsilon stability}
\end{align}\end{definition}
The $\varepsilon$-stability property \eqref{eq: epsilon stability} describes how sensitive the optimizer of the function $\mathbb{E}_{\boldsymbol{\theta}\sim\mathcal{N}(\boldsymbol{\theta}^*,\Sigma)}\big[ g(h(\mathbf{x}|\boldsymbol{\theta}))\big]$ is with respect to the position of the expectation operator, i.e., applied before or after the monotonic function $g(\cdot)$ (see e.g., \cite{morningstar2022pacm, masegosa2020learning} for related definitions). 

\begin{assumption}[Normalized regret conditions] \label{assump:regret_cond}
The normalized regret 
\begin{align}
    \tilde{r}_{n,t+1}(\mathbf{x}|\boldsymbol{\theta}^*) = \frac{\bar{r}_{n,t+1}(\mathbf{x}|\boldsymbol{\theta}^*)}{\sigma_{\boldsymbol{\theta}*}(\mathbf{x})}\label{eq: normalized regret}
\end{align}
is $\varepsilon_1$-sharp around $\boldsymbol{\theta}^*$ as per Definition~\ref{def:1} as well as  $\varepsilon_2$-stable around $\boldsymbol{\theta}^*$ as per Definition~\ref{def:2}. Also, the normalized regret  $\tilde{r}_{n,t}(\mathbf{x}|\boldsymbol{\theta}^*)$ is $L$-Lipschitz continuous. 
\end{assumption}

\begin{assumption}[Convergence of the parameter vector posterior] \label{lemma_3} The posterior $p(\boldsymbol{\theta}|\mathcal{D}_{n-1})$ converges in distribution as 
    \begin{align}
        p(\boldsymbol{\theta}|\mathcal{D}_{n-1}) \overset{d}{\rightarrow} \mathcal{N}\bigg( \boldsymbol{\theta}^*, \frac{1}{n-1} \Sigma_{\boldsymbol{\theta}^*} \bigg), \text{ for } n \rightarrow \infty,
    \end{align}
    for a covariate matrix $\Sigma_{\boldsymbol{\theta}^*}$ satisfying the inequality
    \begin{align}
        \text{Tr}(\Sigma_{\boldsymbol{\theta}^*}) \leq \frac{\xi}{(1-c)^\alpha\cdot T^\alpha}
    \end{align}
    for some constants $0<\xi<\infty$ and  $\alpha > 0$. 
\end{assumption}

Assumption~\ref{lemma_3} can be justified via the Bernstein–von Mises theorem \cite{kleijn2012bernstein}, which guarantees convergence of the posterior $p(\boldsymbol{\theta}|\mathcal{D}_{n-1})$ to a Gaussian distribution $\mathcal{N}(\boldsymbol{\theta}^*, F(\boldsymbol{\theta}^*)^{-1}/n)$, where $F(\boldsymbol{\theta})$ is the Fisher information matrix of the likelihood $p(y|\mathbf{x},\boldsymbol{\theta})$.

In particular, Assumption~\ref{lemma_3} is verified as long as the Fisher information matrix has the property
\begin{align} \label{eq:FIM_bound}
      \text{Tr}(F(\boldsymbol{\theta}^*)^{-1}) =  \mathcal{O}\bigg(\frac{ 1 }{(1-c)^\alpha\cdot T^\alpha}\bigg)
\end{align}
for some $\alpha>0$. For $\alpha=1$, the condition (\ref{eq:FIM_bound}) corresponds to the additivity of Fisher information, which holds for i.i.d. observations $\mathbf{y}=\{f(\mathbf{x}_t)+\epsilon_t\}_{t= \lceil c\cdot T \rceil + 1}^{T}$ in each task. Given that the observations $\mathbf{y}$ are generally correlated, the scaling with $\alpha=1$ in (\ref{eq:FIM_bound}) may not be satisfied. Indeed, as discussed in \cite{fahs2017information}, the Fisher information can be sub-additive, yielding an exponent $\alpha<1$, for positively correlated observations, or super-additive, corresponding to an exponent $\alpha>1$, for negatively correlated observations.

\subsection{Regret Analysis}\label{ssec: regret analysis}
The cumulative expected regret bound of the two-phase MFT-MES protocol described in Sec. \ref{ssec: seq mes unknown}, is bounded in the following theorem.

\begin{theorem}[Cumulative Expected Regret Bound of MFT-MES]  \label{theor}
Under Assumptions~\ref{assump:regret_cond} and \ref{lemma_3}, denoting $\varepsilon=\varepsilon_1+\varepsilon_2$, the cumulative expected regret of the two-phase MFT-MES scheme  satisfies the inequality
\begin{align} \label{eq:theor1}
  &\mathcal{R}_{n,cT}(\mathbf{X}_{n,cT}|\boldsymbol{\theta}^*)\nonumber\\&\leq \left(\nu_{n,t_n^*}+\frac{\varepsilon \cdot L \cdot \xi}{(n-1)\cdot(1-c)^\alpha\cdot T^\alpha}\right)\sqrt{\zeta\cdot c\cdot T\cdot \rho_{n, cT}},
\end{align}
where $\nu_{n,t}=\min_\mathbf{x}\tilde{r}_{n,t}(\mathbf{x}|\boldsymbol{\theta}^*)$ is the minimum normalized expected regret \eqref{eq: normalized regret} at time $t$; $ t_n^* = \arg\max_{t \in \{1,...,c\cdot T\}} \nu_{n,t}$ is the time at which $\nu_{n,t_n}$ is maximized; $0<\xi<\infty$ is a constant defined in Assumption \ref{lemma_3};  $\zeta$ is defined as $\zeta=2/\log(1+\sigma^{-2})$; and the maximum information gain $\rho_{n, cT}$ is 
\begin{align} 
  \rho_{n, cT} = \max_{A \subset \mathcal{X}: |A|=  cT} I_{n}(\mathbf{y}_A; \boldsymbol{f}_A),
\end{align}
which is evaluated as in \cite[Eq. (7)]{srinivas2012tit}.
\end{theorem}

Theorem \ref{theor}, which is proven in Appendix, shows that the cumulative expected regret decreases as the number of tasks $n$ grows large, reflecting the benefits of knowledge transferred across tasks. In particular, devoting a larger fraction of time $(1-c)$ to collect information about the shared parameters $\boldsymbol{\theta}^*$ ensures a faster rate of decrease of the cumulative regret with $n$. On the flip side, choosing an excessively large value of $(1-c)$ can cause a increase in the cumulative regret, as per the second term in (\ref{eq:theor1}), owing to the insufficient time devoted to the exploration of the current objective function. As verified via numerical results in the following section, this result suggests that MFT-MES can be advantageous as soon as $n$ is large enough, as in this case one can choose a value of $c$ near 1, while still guaranteeing a vanishing contribution of the term $(\varepsilon \cdot L \cdot \xi)/((n-1)\cdot(1-c)^\alpha\cdot T^\alpha)$  in (\ref{eq:theor1}).

\section{Experiments}\label{sec: exp}
In this section, we empirically evaluate the performance of the proposed MFT-MES on the synthetic benchmark adopted in \cite{takeno2020multi,moss2021gibbon}, as well as on a real-world application, namely radio resource management for wireless cellular systems \cite{Maggi2021bogp,zhang2023bayesian}. The Supplementary Material contains an additional experiment on a gas emission source term estimation problem \cite{li2017fast}. 

\subsection{Benchmarks}\label{ssec: benchmarks}
We consider the following benchmarks in all experiments: 1) MF-MES \cite{moss2021gibbon}, as described in Algorithm \ref{table: mfmes}; and 2) \emph{Continual MF-MES}, which applies Algorithm \ref{table: mftmes} with $\beta=0$, thus not attempting to transfer knowledge from previous tasks to current task. To the best of our knowledge, Continual MF-MES is also considered for the first time in this work. Unlike MF-MES \cite{moss2021gibbon}, Continual MF-MES share with MFT-MES the use of SVGD for the update of $V$ particles for the parameter vector $\boldsymbol{\theta}$. The $V$ particles $\{\boldsymbol{\theta}_1,...,\boldsymbol{\theta}_V\}$ are carried from one task to the next, implementing a form of continual learning. Unlike MFT-MES, however, the selection of input-fidelity pairs $(\mathbf{x},m)$ aims solely at improving the current task via the MF-MES acquisition function \eqref{eq: af closed}, setting $\beta=0$ in \eqref{eq: final af}.

\begin{figure}[t]
    \centering
    \centerline{\includegraphics[scale=0.4]{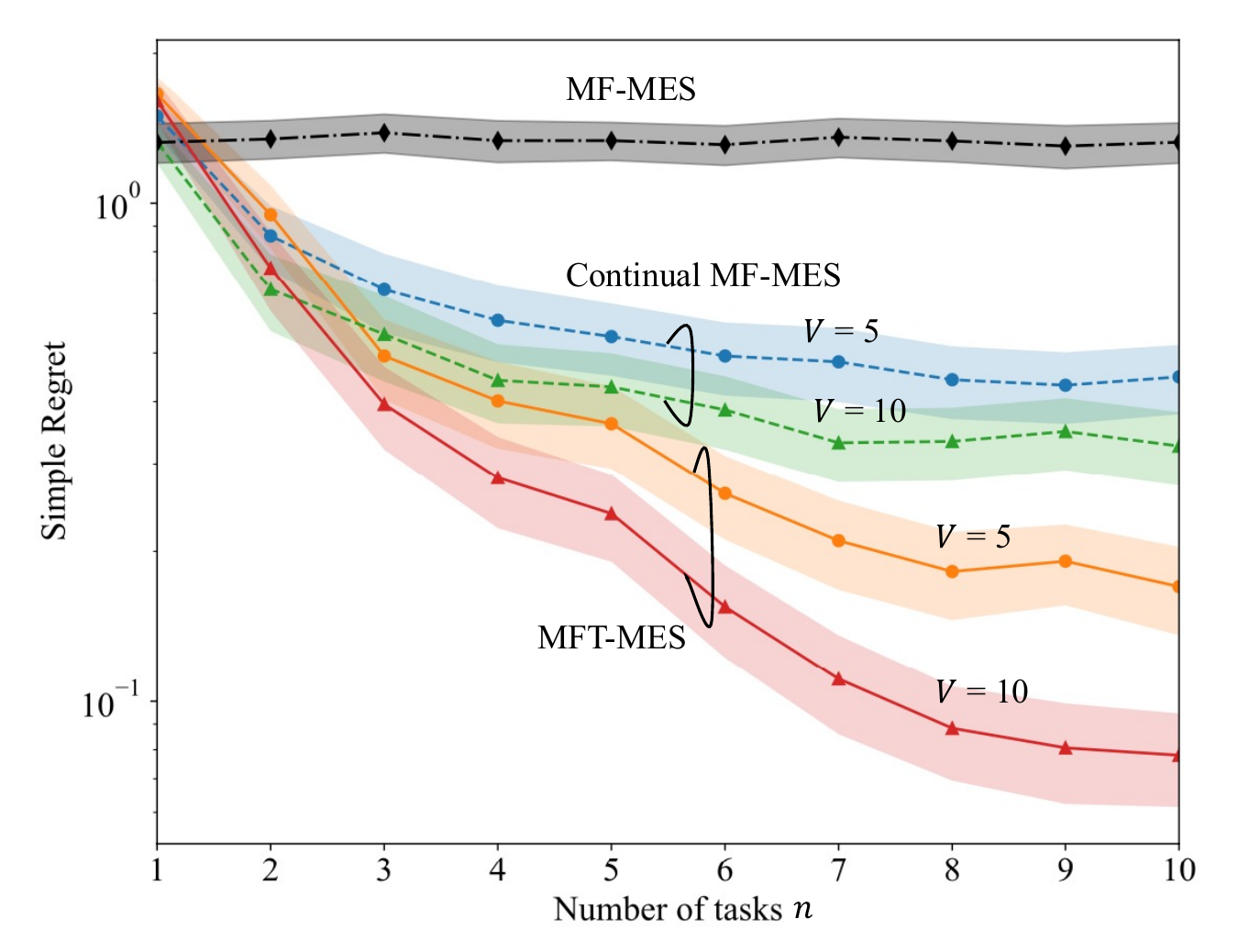}}
    \caption{Synthetic optimization tasks: Simple regret \eqref{eq: simple regret} against the number of tasks, $n$, for MF-MES, Continual MF-MES $(\beta=0)$ with $V=5$ and $V=10$ particles, and MFT-MES ($\beta=1.2$) with $V=5$ and $V=10$ particles.}\label{fig: svgd4}
    \vspace{-0.4cm}
\end{figure}

\subsection{Evaluation and Implementation}\label{ssec: setup}
For all schemes, the parametric mapping $\psi_{\boldsymbol{\theta}}(\cdot)$ in \eqref{eq: sogp para kernel} was instantiated with a fully-connected neural network consisting of three layers, each with $64$ neurons and $tanh$ activation function.  Unless stated otherwise, the MFGP model is initialized with $2d+2$ random evaluations across all fidelity levels before performing BO \cite{moss2021gibbon}, and all the results are averaged over 100 experiments, with figures reporting $90\%$ confidence level. Each experiment corresponds to a random realization of the observation noise signals, random samples of SVGD particles $\{\boldsymbol{\theta}_v\}_{v=1}^V$ at task $n=1$, and random draws of the max-value sets $\mathcal{F}$ in \eqref{eq: af closed}.

For MF-MES, the parameter vector $\boldsymbol{\theta}$ is set as a random sample from the prior distribution $p(\boldsymbol{\theta})$ for each task. The prior distribution is defined as a zero-mean isotropic Gaussian, i.e., $p(\boldsymbol{\theta})=\mathcal{N}(0,\sigma_p^2\mathbf{I})$ with variance $\sigma_p^2=0.5$ across all the experiments. Furthermore, for the SVGD update \eqref{eq: smooth func}, we used a Gaussian kernel $\tilde{k}(\boldsymbol{\theta},\boldsymbol{\theta}')=\exp(-h||\boldsymbol{\theta}-\boldsymbol{\theta}'||^2)$ with bandwidth parameter $h=1/1.326$. We set $S=10$ in \eqref{eq: af closed} and $R=2000$ in Algorithm \ref{table: mftmes}.

\subsection{Synthetic Optimization Tasks}\label{ssec: hartmann}
For the first experiment, we consider synthetic optimization tasks defined by randomly generating Hartmann 6 functions \cite{dixon1978global}, the details of the experimental settings can be found in Supplementary Materials. 

We evaluate the performance of all methods at the end of each task via the \emph{simple regret}, which is defined for a task $n$ as \cite{bubeck2009pure}
\begin{align}
    \text{SR}_{n}=f_n^*-\max_{t=1,...,T_n}f_n^{(M)}(\mathbf{x}_{n,t}).\label{eq: simple regret}
\end{align}
The simple regret \eqref{eq: simple regret} describes the error for the best decision $\mathbf{x}_{n,t}$ made throughout the optimization process.

Fig. \ref{fig: svgd4} shows the simple regret \eqref{eq: simple regret} as a function of the number of tasks $n$ observed so far. For MFT-MES, the weight parameter in \eqref{eq: final af} is set to $\beta=1.2$. Since MF-MES does not attempt to transfer knowledge across tasks, its average performance is constant for all values of $n$. By transferring knowledge across tasks, both Continual MF-MES and MFT-MES can reduce the simple regret as the number of observed tasks $n$ increases, which confirms the theoretical result in Theorem \ref{theor}.

\begin{figure}[t]
  \centering
  \centerline{\includegraphics[width=0.5\textwidth]{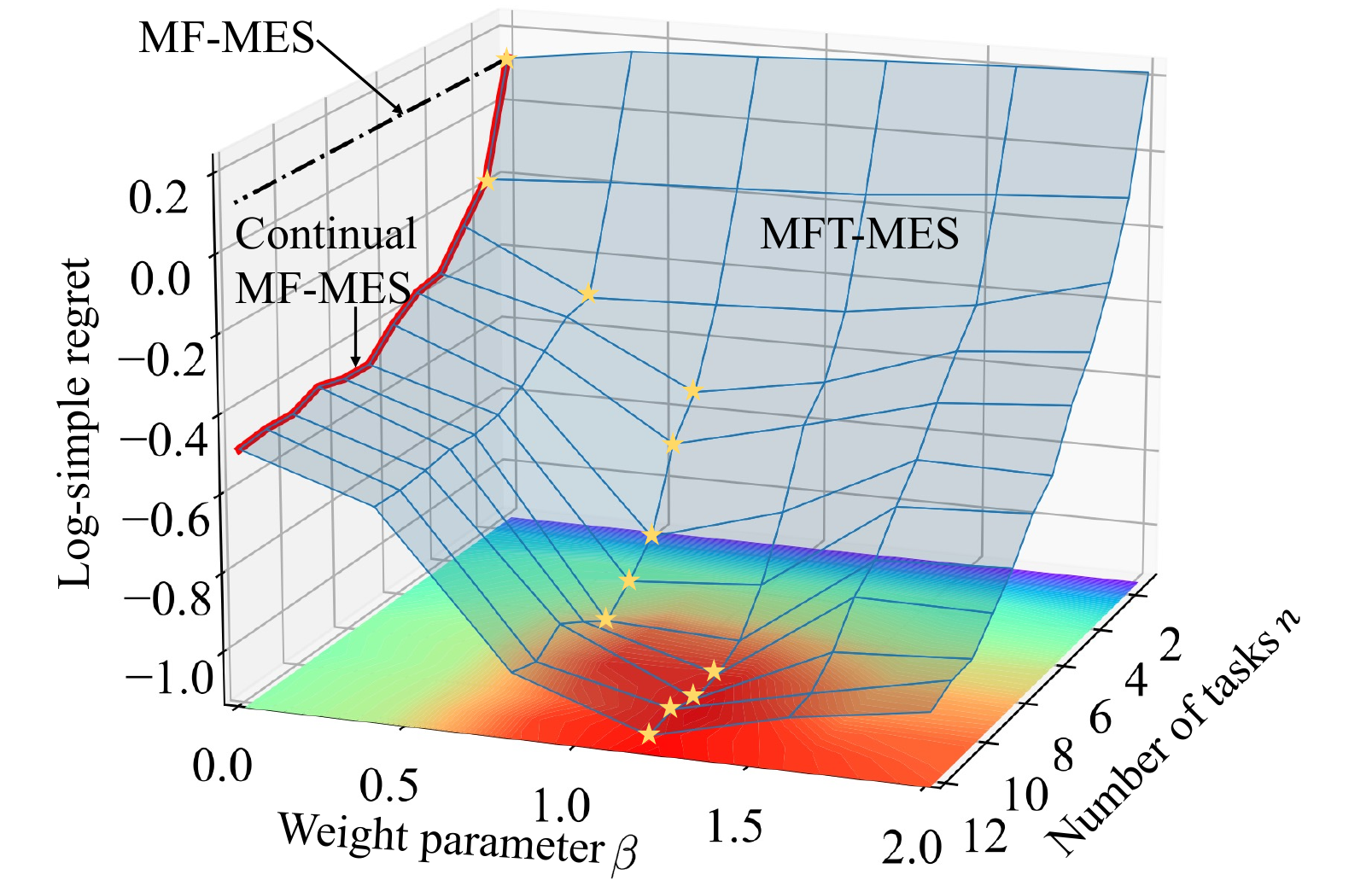}}
  \caption{Synthetic optimization tasks: Log-simple regret against weight parameter $\beta$ and the number of tasks, $n$, for MF-MES (black dash-dotted line), Continual MF-MES (red solid line), and MFT-MES (blue surface). The optimal values of $\beta$ at the corresponding number of tasks $n$ are labeled as gold stars. The number of particles for Continual MF-MES and MFT-MES is set to $V=10$.}\label{fig: impact beta}
\vspace{-0.4cm}
\end{figure}

MFT-MES outperforms Continual MF-MES as soon as the number of tasks, $n$, is sufficiently large, here $n>2$. The advantage of Continual MF-MES for small values of $n$ is due to the fact that MF-MES focuses solely on the current task, while MFT-MES makes decisions also with the goal of improving performance on future tasks. In this sense, the price paid by MFT-MES to collect transferable knowledge is a minor performance degradation for the initial tasks. The benefits of the approach are, however, very significant for later tasks. For instance, at task $n=10$, MFT-MES decreases the simple regret by a factor of three as compared to Continual MF-MES when the number of particles is $V=10$. 

It is also observed that increasing the number of particles $V$, is generally beneficial for both Continual MF-MES and MFT-MES. The performance gain with a larger $V$ can be ascribed to the larger capacity of retaining information about the uncertainty on the optimized parameter vector $\boldsymbol{\theta}$.

Fig. \ref{fig: impact beta} demonstrates the impact of the weight parameter $\beta$ on the simple regret as a function of the number of tasks. Recall that the performance levels of MF-MES, as well as of Continual MF-MES, do not depend on the value of weight parameter $\beta$, which is an internal parameter of the MFT-MES acquisition function \eqref{eq: final af}. The optimal value of $\beta$, which minimizes the simple regret, is marked with a star. It is observed that it is generally preferable to increase the value of $\beta$ as the number of tasks $n$ grows larger. This is because a larger $\beta$ favors the selection of input and fidelity level that focus on the performance of future tasks, and this provident approach is more beneficial for longer time horizons. For example, when the sequence of tasks is short, i.e., when $n\leq2$, the choice $\beta=0$, i.e., Continual MF-MES attains best performance; while the weight parameter $\beta=1.2$ produces best performance given $n=9$ tasks in the sequence.

\begin{figure}[t]
  \centering
  \centerline{\includegraphics[scale=0.39]{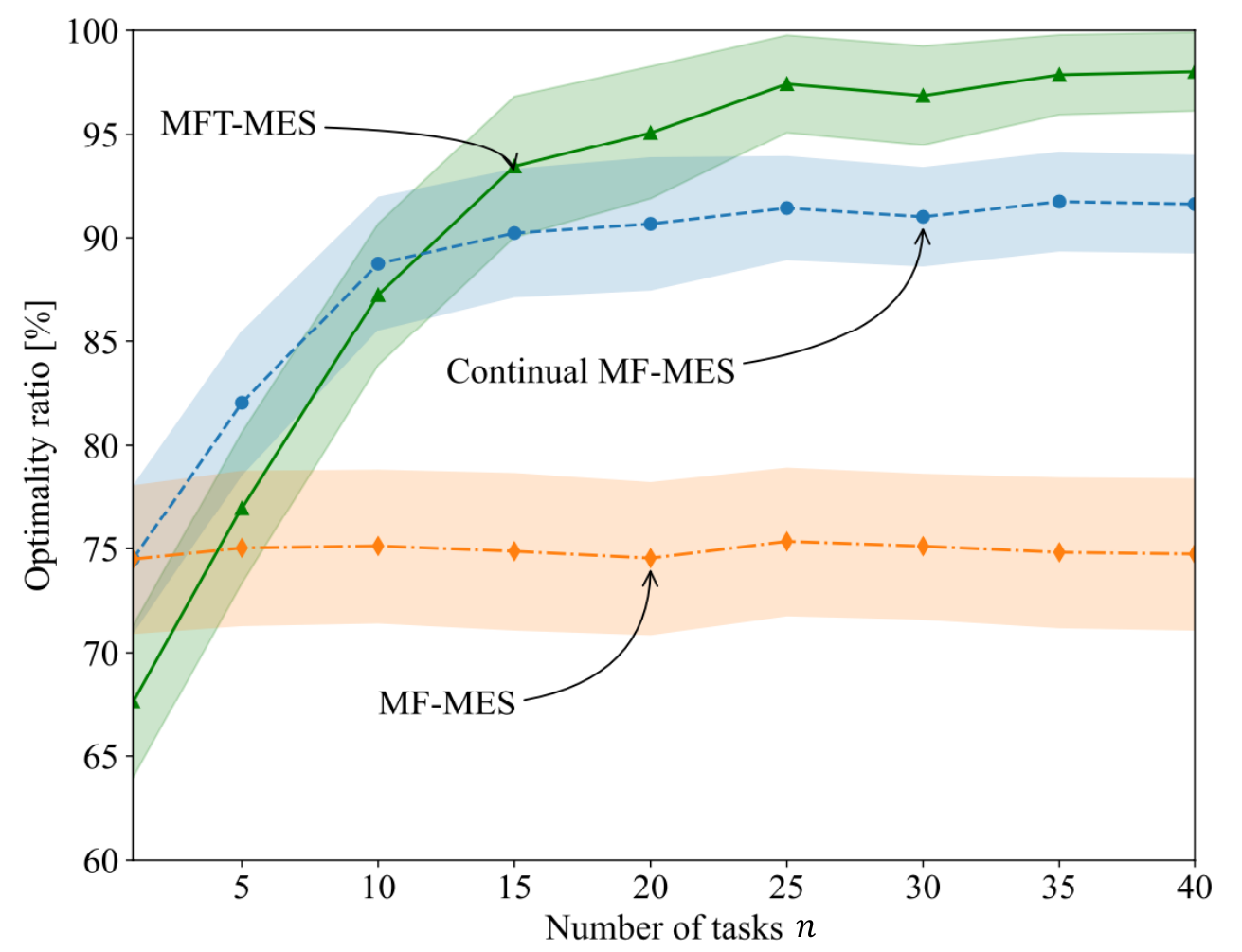}}
  \caption{Radio resource management for wireless systems \cite{zhang2023bayesian}: Optimality ratio \eqref{eq: optimality ratio} against the number of tasks, $n$, for MF-MES, Continual MF-MES ($\beta=0$), and MFT-MES ($\beta=1.6$) with $V=10$ particles.}\label{fig: rrm sr vs task}
\vspace{-0.4cm}
\end{figure}

\subsection{Radio Resource Management}\label{ssec: rrm}
In this section, we study an application to wireless communications presented in \cite{Maggi2021bogp,zhang2023bayesian}. The problem involves optimizing parameters $\mathbf{x}$ that dictate the power allocation strategy of base stations in a cellular system. The vector $\mathbf{x}$ contains two parameters for each base station with the first parameter taking 114 possible values and the second parameter taking 8 possible values. Tasks are generated by randomly deploying users in a given geographical area of radius up to 200 meters, while restricting the users' locations to change by no more than 10 meters for task $n$ as compared to the most recent task $n-1$. 

The objective function $f_n(\mathbf{x})$ is the sum-spectral efficiency at which users transmit to the base stations. Evaluating the sum-spectral efficiency requires averaging out the randomness of the propagation channels, and the simulation costs determines the number of channel samples used to evaluate this average. Accordingly, we set the cost levels as $\lambda^{(1)}=10,\lambda^{(2)}=20,\lambda^{(3)}=50$, and $\lambda^{(4)}=100$, which measure the number of channel samples used at each fidelity level. The total query cost budget is set to $\Lambda=2000$ for every task.

We initialize the MFGP model \eqref{eq: mfgp posterior} with 10 random evaluations across all fidelity levels, and the observation noise variance is $\sigma^2=0.83$. In a manner consistent with \cite{zhang2023bayesian}, the performance of each method is measured by the \emph{optimality ratio}
\begin{align}
    \max\limits_{t=1,..,T_n}\frac{f^{(M)}_n(\mathbf{x}_{n,t})}{f_n^*},\label{eq: optimality ratio}
\end{align}
which evaluates the best function of the optimal value $f_n^*$ attained during the optimization process. Finally, we set $V=10$ particles for Continual MF-MES and MFT-MES. All other experimental settings are kept the same as in Sec. \ref{ssec: setup}.

\begin{figure}[t]
  \centering
  \centerline{\includegraphics[scale=0.39]{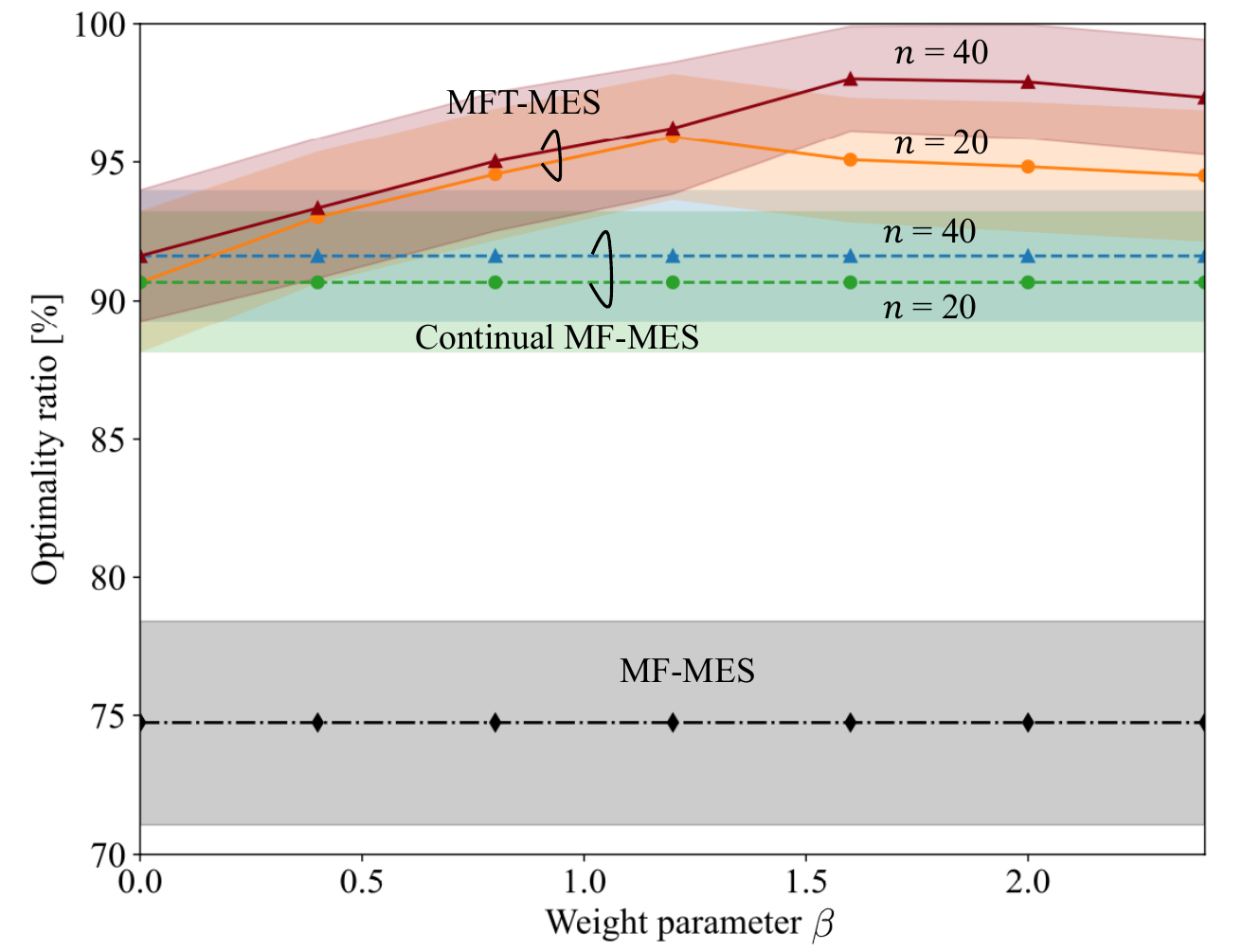}}
  \caption{Radio resource management for wireless systems \cite{zhang2023bayesian}: Optimality ratio \eqref{eq: optimality ratio} against weight parameter $\beta$, for MF-MES, Continual MF-MES with $n=20$ and $n=40$ tasks observed, and MFT-MES with $n=20$ and $n=40$ tasks observed. The number of particles for Continual MF-MES and MFT-MES is set to $V=10$.}\label{fig: rrm sr vs beta}
\vspace{-0.4cm}
\end{figure}

In Fig. \ref{fig: rrm sr vs task}, we set the weight parameter $\beta=1.6$ for MFT-MES and plot the optimality ratio \eqref{eq: optimality ratio} as a function of the number of tasks, $n$. Confirming the discussions in Sec. \ref{ssec: hartmann} and Theorem \ref{theor}, the performance of MF-MES is limited by the lack of information transfer across tasks. In contrast, the performance of both Continual MF-MES and MFT-MES benefits from information transfer. Furthermore, MFT-MES outperforms Continual MF-MES after processing 12 tasks. Through a judicious choice of input and fidelity levels targeting the shared parameters $\boldsymbol{\theta}$, at the end of 40-th task, MFT-MES provides, approximately, an $7\%$ gain in terms of optimality ratio over Continual MF-MES, and a $23\%$ gain over MF-MES.

The impact of the weight parameter $\beta$ used by MFT-MES on the performance evaluated at the 20-th and 40-th task is illustrated in Fig. \ref{fig: rrm sr vs beta}. MFT-MES with any weight parameter $\beta>0$ outperforms all other schemes given the same number of tasks observed so far. The best performance of MFT-MES is obtained for $\beta=1.2$ at $n=20$ tasks, and for $\beta=1.6$ at $n=40$ tasks. Therefore, as discussed in Sec. \ref{ssec: hartmann}, a larger value of weight parameter $\beta$ is preferable as the number of tasks $n$ increases. Moreover, the performance of MFT-MES is quite robust to the choice of $\beta$. For instance, selecting larger weights $\beta$, up to $\beta=2.4$, is seen to yield a mild performance degradation as compared to the best settings of weight parameters $\beta=1.2$ for $n=20$ and $\beta=1.6$ for $n=40$.

\section{Conclusion}\label{sec: conclusion}
In this work, we have have  introduced MFT-MES, a novel information-theoretic acquisition function that balances the need to acquire information about the current task with goal of collecting information transferable to future tasks. The key mechanism underlying MFT-MES involves modeling transferable knowledge across tasks via shared GP model parameters, which are integrated into the acquisition function design and updated following Bayesian principles. From theoretical analysis and empirical results, we have demonstrated that the proposed MFT-MES can obtain performance gains as large as an order of magnitude in terms of simple regret as compared to the state-of-art scheme that do not cater to the acquisition of transferable knowledge.

Future work may address theoretical performance guarantees in the forms of regret bounds to explain aspects such as the dependence of the optimal weight parameter $\beta$ used by MFT-MES as a function of the number of tasks and total query cost budget. Furthermore, it would be interesting to investigate the extensions to multi-objective multi-fidelity optimization problems \cite{li2024evolutionary} or safe optimization problems with unknown constraints \cite{zhang2024bayesian}; the scalability to higher dimensions of the search space \cite{maddox2021bayesian}; and the potential gains from incorporating generative models  \cite{liang2022evolutionary}.

\newpage

\appendix

\subsection{Proof of Theorem \ref{theor}}\label{app: theorem proof} 

We start the proof of Theorem \ref{theor} by deriving an upper bound over the discrepancy in the normalized expected regret with known parameter vector $\boldsymbol{\theta}^*$.  Recall that the first set of iterations is devoted for the current task, i.e., the optimizer selects  $\mathbf{x}_{n,t+1} = \arg\max_{\mathbf{x}}\mathbb{E}_{\boldsymbol{\theta} \sim p(\boldsymbol{\theta}|\mathcal{D}_{n-1}, \mathcal{D}_{n,t})}[ \bar{\alpha}_{\boldsymbol{\theta}}^\text{MES}(\mathbf{x})]$ for $t+1\leq c\cdot T$. Under Assumption~\ref{assump:regret_cond}, i.e., $\tilde{r}_{n,t}(\mathbf{x}|\boldsymbol{\theta})$ is $\epsilon_1$-sharp as per Definition~\ref{def:1}  as well as  $\epsilon_2$-stable as per Definition~\ref{def:2}, and Assumption~\ref{lemma_3},
we have 
\begin{align}
    &\big|\big|\mathbf{x}_{n,t} - \arg\min_{\mathbf{x}\in\mathcal{X}} \tilde{r}_{n,t}(\mathbf{x}|\boldsymbol{\theta}^*)\big|\big|\nonumber\\& \leq  (\varepsilon_1+\varepsilon_2)\frac{\xi}{(n-1)\cdot(1-c)^\alpha\cdot T^\alpha}, \text{ for sufficiently large $n$}
\end{align}
from the triangle inequality.
Furthermore, by assuming the $L$-Lipschitz continuity of $\tilde{r}_{n,t}(\mathbf{x}|\boldsymbol{\theta}^*)$, we have
\begin{align} \label{eq:delta_def}
    \Delta_{n,t}&=\tilde{r}_{n,t}(\mathbf{x}_{n,t}|\boldsymbol{\theta}^*) - \min_{\mathbf{x}\in\mathcal{X}} \tilde{r}_{n,t}(\mathbf{x}|\boldsymbol{\theta}^*)\nonumber\\&\leq L\cdot \big|\big| \mathbf{x}_{n,t} - \arg\min_\mathbf{x} \tilde{r}_{n,t}(\mathbf{x}|\boldsymbol{\theta}^*) \big|\big| \nonumber\\&\leq \frac{\varepsilon \cdot L \cdot \xi}{(n-1)\cdot(1-c)^\alpha\cdot T^\alpha},
\end{align}
where  we have defined $\varepsilon=\varepsilon_1+\varepsilon_2$.

The expected regret of round $t$ at task $n$ is 
\begin{align} \label{eq:pf_1}
  \bar{r}_{n,t}(\mathbf{x}_{n,t}|\boldsymbol{\theta}^*)&= \hat{m}_{n,t}(\boldsymbol{\theta}^*) - \mu_{\boldsymbol{\theta}^*}(\mathbf{x}_t)\nonumber\\&=\frac{\hat{m}_{n,t}(\boldsymbol{\theta}^*)- \mu_{\boldsymbol{\theta}^*}(\mathbf{x}_t)}{\sigma_{\boldsymbol{\theta}^*}(\mathbf{x}_t)} \cdot \sigma_{\boldsymbol{\theta}^*}(\mathbf{x}_t)   \nonumber\\& = \tilde{r}_{n,t}(\mathbf{x}_t|\boldsymbol{\theta}^*)\cdot \sigma_{\boldsymbol{\theta}^*}(\mathbf{x}_t) \nonumber\\& =(\nu_{n,t} + \Delta_{n,t} )\cdot  \sigma_{\boldsymbol{\theta}^*}(\mathbf{x}_t),
\end{align}
where the last equality is obtained from \eqref{eq:delta_def}.


Therefore, based on \eqref{eq:pf_1}, we have 
\begin{align} \label{eq:pf_2}
  \mathcal{R}_{n,cT}(\mathbf{X}_{n,cT}|\boldsymbol{\theta}^*)&= \sum_{t=1}^{c \cdot T} (\nu_{n,t}+ \Delta_{n,t})\cdot \sigma_{\boldsymbol{\theta}^*}(\mathbf{x}_t)\nonumber\\&
  \leq  (\nu_{n,t_n^*} + \Delta_{n,t_n^*} )\sum_{t=1}^{c \cdot T} \sigma_{\boldsymbol{\theta}^*}(\mathbf{x}_t)\nonumber\\&\leq 
 (\nu_{n,t_n^*} + \Delta_{n,t_n^*}) \sqrt{c\cdot T \sum_{t=1}^{c \cdot T} \sigma_{\boldsymbol{\theta}^*}^2(\mathbf{x}_t)} \nonumber\\&\leq 
  (\nu_{n,t_n^*} + \Delta_{n,t_n^*} )\sqrt{\frac{2c\cdot T\cdot \rho_{n,cT}}{\log(1+\sigma^{-2})}},
\end{align}
where the second inequality is obtained directly from the definition of $t_n^*$; the third inequality comes from Cauchy-Schwarz inequality; and the last inequality is a direct consequence of \cite[Lemma 5.3]{srinivas2012tit} and \cite[proof of Theorem 3.2]{wang2017max}.

\subsection{Discussion on Theorem \ref{theor}}\label{app: theorem 1 plot}
The theoretical analysis presented in Sec. \ref{sec: theorems} applies the following simplifications to MFT-MES:
\begin{enumerate}
    \item \textbf{Separate per-task optimization and acquisition of shared knowledge:} Each optimization step focuses either on enhancing per-task performance or on reducing the uncertainty on the shared parameter vector $\boldsymbol{\theta}$.
    \item \textbf{Myopic shared knowledge acquisition:} The acquisition function used in the steps devoted to the reduction of uncertainty on the parameter vector $\boldsymbol{\theta}$ relies on a batch-wise mutual information criterion \eqref{eq:batch_MES} that disregards observations from previous tasks.
    \item \textbf{Two-phase protocol:} The iterations are divided into two separate phases, with the first fraction $c$ of iterations used for per-task optimization, and the remaining fraction $1-c$ of iterations devoted to acquire shared knowledge.
\end{enumerate}
The performance of the resulting simplified scheme is expected to be worse than the proposed MFT-MES algorithm, thus intuitively providing an upper bound on the regret performance of MFT-MES.

In order to gain insights into the gap between the performance of the analyzed simplified scheme and of MFT-MES, this appendix evaluates the performance of the following successive simplifications of MFT-MES through a numerical experiment:
\begin{enumerate}
    \item \textbf{Separation-based MFT-MES:} This scheme applies only the first simplification, namely separate per-task optimization and acquisition of shared knowledge. Accordingly, the next candidate solution is selected as
    \begin{align}
     \mathbf{x}_{n,t+1}=\begin{cases}
    \arg\max\limits_{\mathbf{x}\in\mathcal{X}}\mathbbm{E}_{p(\boldsymbol{\theta}|\mathcal{D}_{n-1},\mathcal{D}_{n,t})}[\alpha_{\boldsymbol{\theta}}^{\text{MF-MES}}(\mathbf{x},M)]& \\\quad\text{with probability $c$}, \\
    \arg\max\limits_{\mathbf{x}\in\mathcal{X}}\frac{\Delta\alpha^{\text{MFT-MES}}(\mathbf{x},M)}{\lambda^{(M)}}& \\\quad\text{with probability $1-c$}.\label{eq: quantizer}
    \end{cases}
    \end{align}
    Accordingly, the algorithm opts independently at each iteration to either enhance the current task performance or acquire transferable knowledge.
    \item \textbf{Separation-based Myopic MFT-MES:} This scheme applies the first two simplifications, selecting the next candidate solution as
    \begin{align}
     \mathbf{x}_{n,t+1}=\begin{cases}
    \arg\max\limits_{\mathbf{x}\in\mathcal{X}}\mathbbm{E}_{p(\boldsymbol{\theta}|\mathcal{D}_{n-1},\mathcal{D}_{n,t})}[\alpha_{\boldsymbol{\theta}}^{\text{MF-MES}}(\mathbf{x},M)]&\\ \quad \text{with probability $c$}, \\
    \arg\max\limits_{\mathbf{x}\in\mathbf{X}} I(\boldsymbol{\theta};y|\mathbf{x})\text{ for $\mathbf{X}$ in \eqref{eq:batch_MES}}& \\\quad\text{with probability $1-c$}.\label{eq: prob myopic}
    \end{cases}
    \end{align}
    This implies that the algorithm opts independently at each iteration to either focus on current task optimization or to acquire shared knowledge by adopting the batch-wise mutual information criterion \eqref{eq:batch_MES}.
    \item \textbf{Two-phase MFT-MES (Two-phase):} This scheme applies all the simplifications, corresponding to the MFT-MES variant studied in Sec. \ref{ssec: seq mes unknown} for theoretical analysis. 
\end{enumerate}

\begin{figure}[t]
    \centering
    \centerline{\includegraphics[scale=0.4]{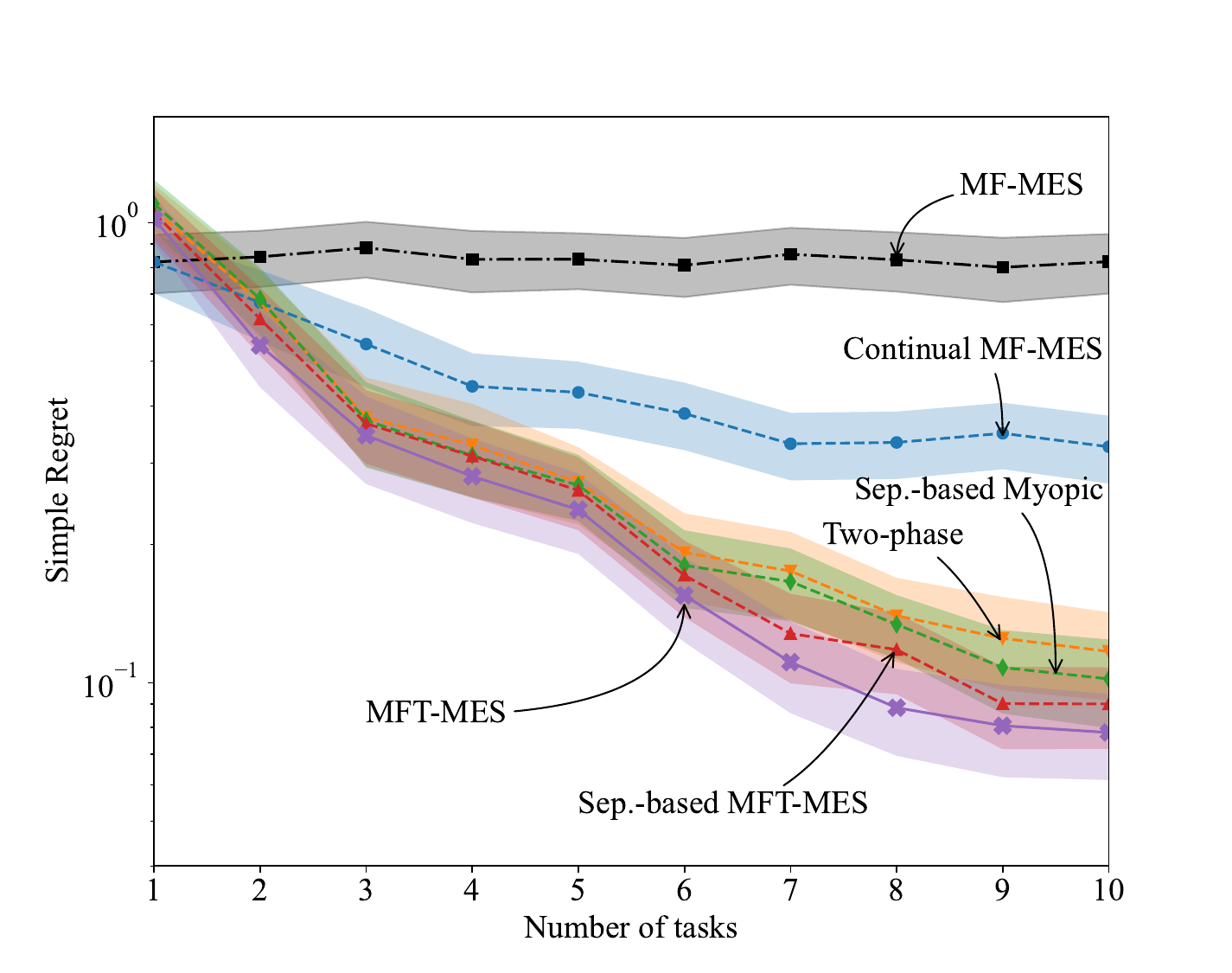}}
    \caption{Synthetic optimization tasks: Simple regret \eqref{eq: simple regret} against the number of tasks, $n$, for MF-MES (black dot-dashed line), Continual MF-MES (blue dashed line), MFT-MES with $\beta=1.5$ (purple solid line), along with three simplified versions of MFT-MES (see Appendix B): Separation-based MFT-MES (red dashed line), Separation-based Myopic MFT-MES (green dashed line) and Two-phase MFT-MES (orange dashed line). The number of SVGD particles is set to $V = 10$.}\label{fig: verify theorem}
\end{figure}

Fig. \ref{fig: verify theorem} shows the simple regret \eqref{eq: simple regret} as a function of the number of tasks $n$ observed so far. The weight parameter in \eqref{eq: final af} for MFT-MES is set to $\beta=1.5$, the fraction parameter for the simplified schemes is set to $c=0.6$, and other numerical settings are kept the same as in Sec. \ref{ssec: setup}. As illustrated in Fig. \ref{fig: verify theorem}, each successive simplification deteriorates the performance of MFT-MES. However, the performance gaps are fairly marginal. Furthermore, all the simplified variants of MFT-MES always outperform the benchmarks MF-MES and Continual MF-MES, as long as the number of tasks is larger than $n=2$. This result provides empirical evidence that our theoretical analysis offers relevant insights into the benefits of MFT-MES by studying a simplified scheme that retains the main features of MFT-MES. Further experimental results can be found in the supplementary material.

\bibliographystyle{IEEEtran}

\bibliography{refer}


\end{document}


%
\title{Supplementary Material of ``Multi-Fidelity Bayesian Optimization With Across-Task Transferable Max-Value Entropy Search"}

\author{\IEEEauthorblockN{Yunchuan Zhang, Sangwoo Park, and Osvaldo Simeone }
}


%


\maketitle



%
\IEEEpeerreviewmaketitle

\section{Single-Task MES with Known Parameter Vector}\label{sec: supporting lemmas}

In \cite{wang2017max}, the single-task MES acquisition function is defined as
\begin{align}\tag{61}
    \alpha_{\boldsymbol{\theta}^*}^\text{MES}(\mathbf{x}) = I(f_n^*;y_n|\mathbf{x},\boldsymbol{\theta}^*, \mathcal{D}_{n,t}),\label{eq: mes acq func}
\end{align}
and the expected regret (40) is derived for a fixed and known parameter vector $\boldsymbol{\theta}^*$.

A related acquisition function was studied in \cite{wang2016optimization}, which is given by 
\begin{align}\tag{62} \label{eq:expected_MSE}
    \bar{\alpha}_{\boldsymbol{\theta}^*}^\text{MES}(\mathbf{x}) =  g\bigg(\frac{\hat{m}_{n,t}(\boldsymbol{\theta}^*)  - \mu_{\boldsymbol{\theta}^*}(\mathbf{x}) }{\sigma_{\boldsymbol{\theta}^*}(\mathbf{x})}\bigg),
\end{align} 
where $\hat{m}_{n,t}(\boldsymbol{\theta}^*) = \mathbb{E}_{f_n(\mathbf{x})}[ f_n^* ]$ is the expected maximum value of the function. Note that the acquisition functions (\ref{eq: mes acq func}) and (\ref{eq:expected_MSE}) coincide when approximating the expectation in (\ref{eq:expected_MSE}) with a single sample $f_n^*=\max_{\mathbf{x}}f_n(\mathbf{x}) \text{with} f_n(\mathbf{x})\sim\mathcal{N}(\mu_{\boldsymbol{\theta}^*}(\mathbf{x}), \sigma^2_{\boldsymbol{\theta}^*}(\mathbf{x}))$.

Using the monotonicity property of function $g(\cdot)$ in (45), the maximizer of the acquisition function (\ref{eq:expected_MSE}) is given by 
\begin{align} \label{eq:x_nt_expected_regret}
    \mathbf{x}_{n,t+1}&=\arg\max\limits_{\mathbf{x}\in\mathcal{X}} \bar{\alpha}_{\boldsymbol{\theta}^*}^\text{MES}(\mathbf{x})  \nonumber\\&= \arg\min\limits_{\mathbf{x}\in\mathcal{X}} \bigg\{\frac{\hat{m}_{n,t}(\boldsymbol{\theta}^*)-\mu_{\boldsymbol{\theta}^*}(\mathbf{x}) }{\sigma_{\boldsymbol{\theta}^*}(\mathbf{x})}\bigg\} \nonumber\\&= \arg\min\limits_{\mathbf{x}\in\mathcal{X}} \bigg\{\frac{\bar{r}_{n,t+1}(\mathbf{x}|\boldsymbol{\theta}^*)}{\sigma_{\boldsymbol{\theta}^*}(\mathbf{x})}\bigg\}.\tag{63}
\end{align}
Hereby, the acquisition function (\ref{eq:expected_MSE}) chooses the next candidate solution $\mathbf{x}_{n,t+1}$ so as to maximize  the normalized expected regret $\tilde{r}_{n,t+1}(\mathbf{x}|\boldsymbol{\theta}^*)$ in (47) which is non-negative. Using the candidate solutions \eqref{eq:x_nt_expected_regret} at each iteration, references \cite{wang2016optimization, wang2017max} demonstrated the following expected regret.

\begin{manuallemma}{1}[Expected regret bound for MES with known parameter vector $\boldsymbol{\theta}^*$    {\protect{\cite[Theorem 3.2]{wang2017max}}}] \label{lemma_2} For a bounded kernel satisfying $k_{\boldsymbol{\theta}}(\mathbf{x},\mathbf{x}')\leq 1$ and a fixed known parameter vector $\boldsymbol{\theta}^*$, MES with acquisition function (\ref{eq:x_nt_expected_regret}) attains the cumulative expected regret 
\begin{align} \label{eq:MES_upper}
  \mathcal{R}_{n,T}(\mathbf{X}_{n,T}|\boldsymbol{\theta}^*) &\leq \underbrace{\nu_{n,t_n^*}  \sqrt{\zeta\cdot T \cdot \rho_{n,  T}}}_{= \mathcal{R}^\text{u}_{n,T}(\mathbf{x}_{n,1:T}|\theta^*)},\tag{64}
\end{align}
where $\zeta=2/\log(1+\sigma^{-2})$ and $\nu_{n,t}=\min_\mathbf{x}\tilde{r}_{n,t}(\mathbf{x}|\boldsymbol{\theta}^*)$ is the minimum normalized expected regret at time $t$;  $t_n^* = \arg\max_{t \in \{1,..., T\}} \nu_{n,t}$ is the time at which the minimum normalized regret $\nu_{n,t}$ is maximized; and  we have denoted as
\begin{align} \label{eq:MI_gain}
  \rho_{n, T} = \max_{A \subset \mathcal{X}: |A|=  T} I_{n}(\mathbf{y}_A; \boldsymbol{f}_A)\tag{65}
\end{align}
the maximum information gain, in which the mutual information between the two values of the objective $\boldsymbol{f}_A$ evaluated at candidates $A=\{\tilde{\mathbf{x}}_t\}_{t=1}^{T}$, i.e., $\boldsymbol{f}_A =\{f_n(\tilde{\mathbf{x}}_{t})\}_{t=1}^T$,  and the corresponding observations $\mathbf{y}_A = \{\tilde{y}_{t} \}_{t=1}^T$ with $\tilde{y}_t = f_n(\tilde{\mathbf{x}}_{t}) + \epsilon_{n,t}$, is evaluated as
\begin{align}
    I_{n}(\mathbf{y}_A;\boldsymbol{f}_A)=\frac{1}{2}\sum_{t=1}^{T}\log(1 + \sigma^{-2}\sigma_{\boldsymbol{\theta}^*}^2(\tilde{\mathbf{x}}_t|\tilde{\mathcal{D}}_{n,t-1})),\tag{66}
\end{align}
where $\tilde{\mathcal{D}}_{n,t} = \{ (\tilde{\mathbf{x}}_{t'}, \tilde{y}_{t'}) \}_{t'=1,...,t}$ is the hypothetical dataset for the maximum information gain.
\end{manuallemma}

\renewcommand{\thefigure}{8}
\begin{figure}[t]
    \centering
    \centerline{\includegraphics[scale=0.37]{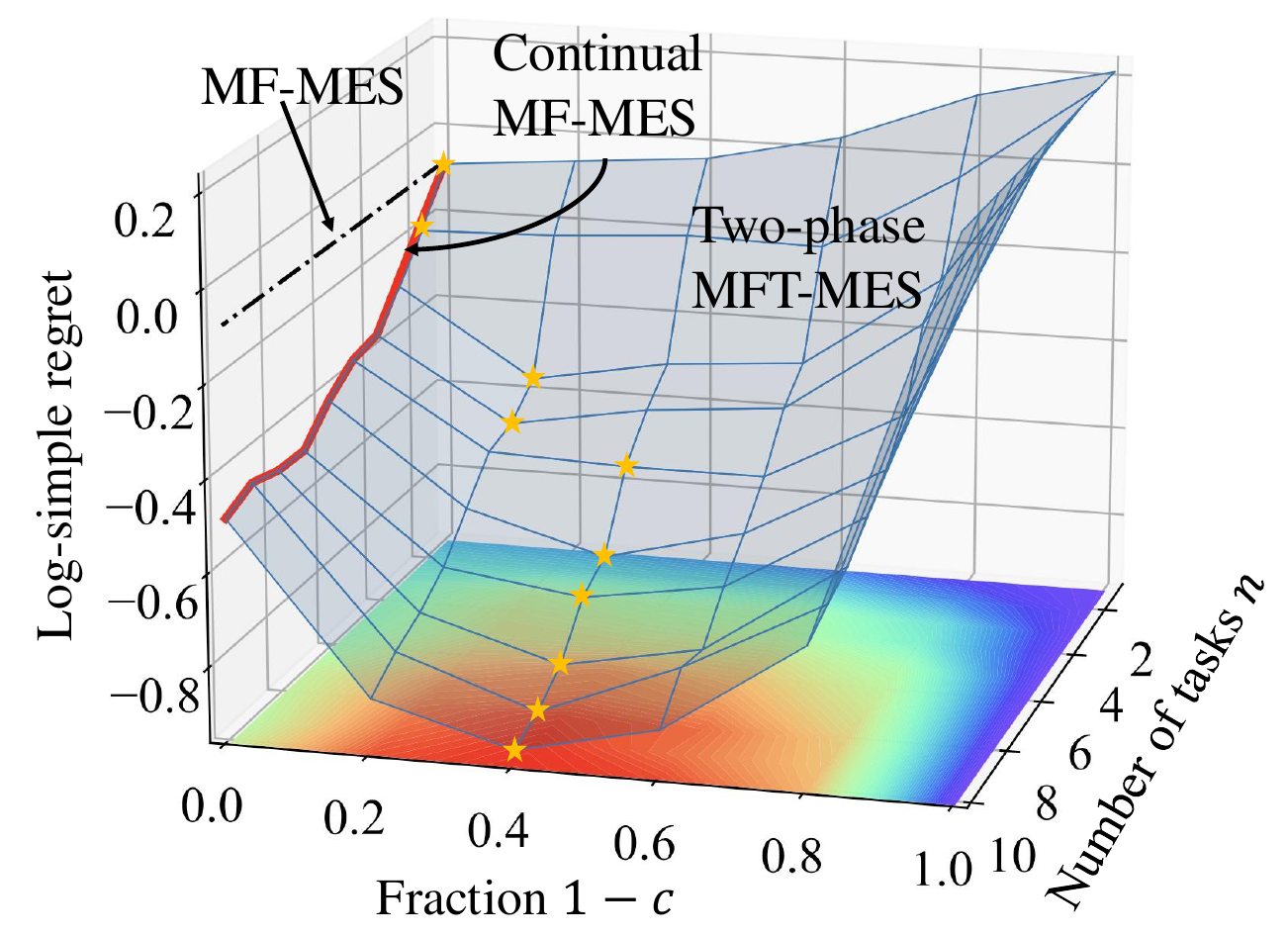}}
    \caption{Synthetic optimization tasks: Simple regret (53) against the fraction $1-c$ and the number of tasks $n$, for MF-MES (black dot-dashed line), Continual MF-MES (red solid line) and Two-phase MFT-MES (blue surface). The optimal values of $1-c$ at the corresponding number of tasks $n$ are labeled as gold stars. The number of SVGD particles is set to $V = 10$.}\label{fig: 3d plot 2}
\end{figure}

\section{Experimental Details for Synthetic Optimization tasks}\label{sec: synthetic}

We now provide detailed information about generating synthetic optimization tasks based on Hartmann 6 function.

The input domain is defined as $\mathcal{X}=[0,1]^6$, and the objective value $f^{(m)}_n(\mathbf{x})$ for a task $n$ at fidelity level $m$ is obtained as \cite{moss2021gibbon}

\begin{align}
    &f_n^{(m)}(\mathbf{x})=-\sum_{i=1}^4 a_{i,m}\exp\bigg(-\sum_{j=1}^6 \Delta_{i,j,n} A_{i,j}(x_j-P_{i,j})^2\bigg), \tag{67}\label{eq: hartmann6}
\end{align}
where $a_{i,m}$, $A_{i,j}$ and $P_{i,j}$ are the $(i,m)$-th, $(i,j)$-th, and $(i,j)$-th entries, respectively, of matrices
\begin{align}
&\mathbf{a}=\begin{pmatrix}
1 & 1.01 & 1.02 & 1.03\\
1.2 & 1.19 & 1.18 & 1.17\\
3 & 2.9 & 2.8 & 2.7\\
3.2 & 3.3 & 3.4 & 3.5
\end{pmatrix}, \nonumber\\&\mathbf{A}=\begin{pmatrix}
10 & 3 & 17 & 3.5 & 1.7 & 8\\
0.05 & 10 & 17 & 0.1 & 8 & 14\\
3 & 3.5 & 1.7 & 10 & 17 & 8\\
17 & 8 & 0.05 & 10 & 0.1 & 14
\end{pmatrix},\nonumber \\ &\text{and}\,\,\mathbf{P}=10^{-4}\begin{pmatrix}
1312 & 1696 & 5569 & 124 & 8283 & 5886\\
2329 & 4135 & 8307 & 3736 & 1004 & 9991\\
2348 & 1451 & 3522 & 2883 & 3047 & 6650\\
4047 & 8828 & 8732 & 5743 & 1091 & 381
\end{pmatrix}.\tag{68}
\end{align}
Optimization tasks differ due to the parameter $\Delta_{i,j,n}$ in \eqref{eq: hartmann6} which are generated in an i.i.d. manner from the uniform distribution $\mathcal{U}(0.8,1.2)$. We set the cost levels as $\lambda^{(1)}=10,\lambda^{(2)}=15,\lambda^{(3)}=20,$ and $\lambda^{(4)}=25$, the total query cost budget in constraint (9) to $\Lambda=500$ for all tasks, and choose the observation noise variance in (6) as $\sigma^2=0.1$. 

\begin{figure}[t]
  \centering
  \centerline{\includegraphics[scale=0.39]{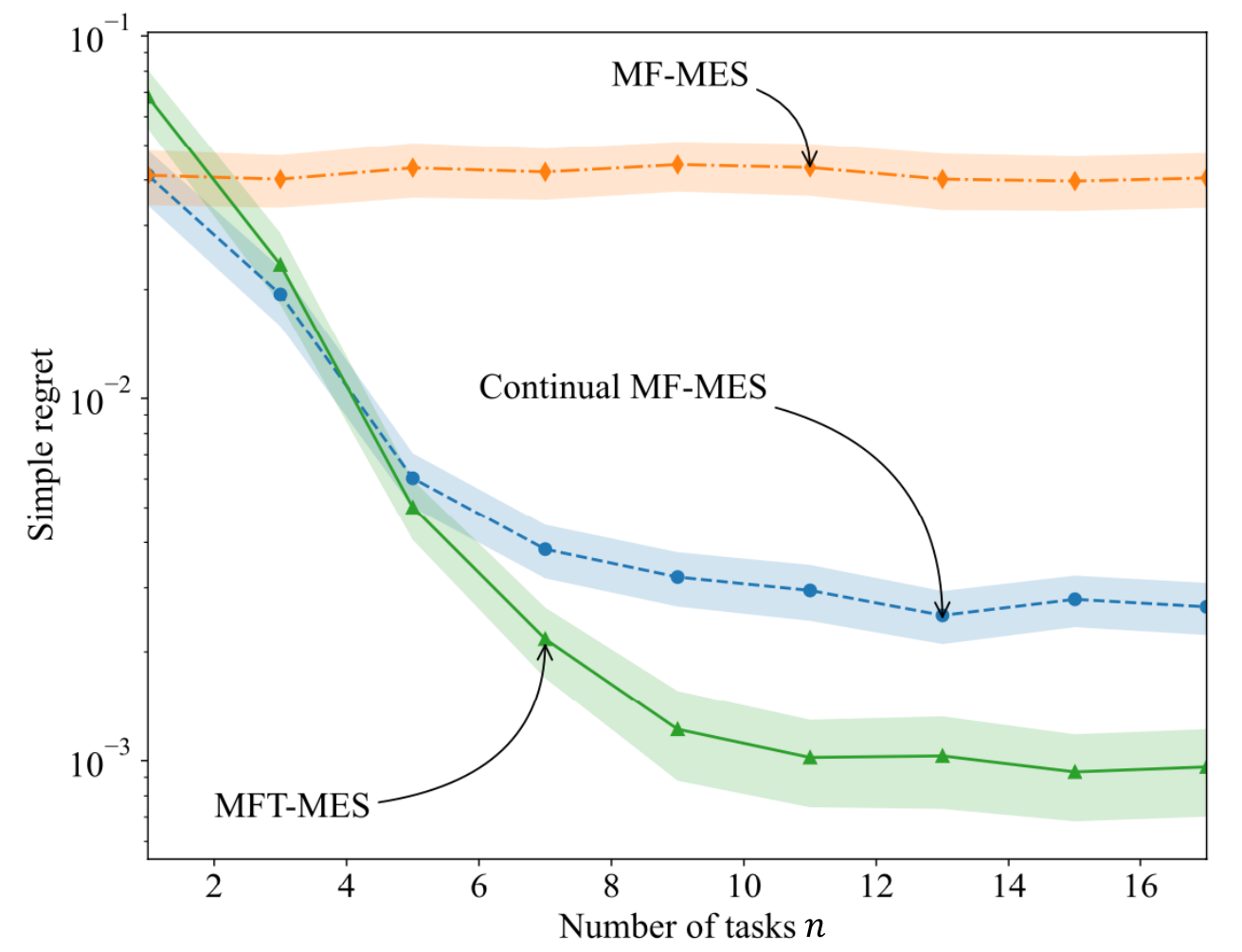}}
  \figuretag{9}
  \caption{Gas emission source term estimation: Simple regret (53) against the number of tasks, $n$, for MF-MES, Continual MF-MES $(\beta=0)$, and MFT-MES ($\beta=1.5$) with $V=10$ particles.}\label{fig: gas sr vs n}
\end{figure}

\section{Further Discussion on Theorem 1}\label{sec: added discussion theorem 1}
The results in Fig. 7 hinges on a choice of parameter $c$ that is well aligned with the behavior of the original MFT-MES scheme, which depends on the selection of the regularization coefficient $\beta$. To bring further evidence about the role of parameter $c$ and its relationship with $\beta$, we report in Fig. \ref{fig: 3d plot 2} the simple regret (53) as a function of the number of tasks and of the parameter $1-c$. It is observed that it is generally preferable to increase the value of $1-c$ as the number of tasks $n$ grows larger, since a larger value of $1-c$ allows the Two-phase MFT-MES scheme to focus more on the performance for future tasks. The optimal values of $1-c$ as well as overall performance of Two-phase MFT-MES reproduce closely the same trends as for the original MFT-MES scheme in Fig. 4. This offers more empirical evidence that simplified MFT-MES still retains the main features of MFT-MES.

\section{Additional Experiments on Gas Emission Source Term Estimation}\label{sec: gas}

We consider an application to the reverse problem formulation of gas emission source term estimation introduced in \cite{li2017fast}. The problem aims to optimize a decision vector $\mathbf{x}$ that identifies the characteristics of the gas emission point source based on the Pasquill-Gifford dispersion model \cite{crowl2001chemical}. The feasible input domain of the vector is defined as $\mathbf{x}\in[10,5000]\times[-500,500]^2\times[0,10]\subset\mathbbm{R}^4$, with the first parameter being the source emission rate and the rest of the parameters describing the location of the emission source. Tasks are distinguished by the different locations of the sensors used to measure the concentration of emissions.

The objective function $f_n(\mathbf{x})$ is defined as the sum of the squared errors between the concentration measured at the sensors and the concentration calculated by the dispersion model given parameters $\mathbf{x}$. The fidelity of the evaluation of the objective function $f_n(\mathbf{x})$ depends on the atmospheric conditions, which can be classified into $M=6$ fidelity levels controlled by dispersion coefficients as in \cite{li2017fast}. We set the cost levels as $\lambda^{(m)}=10+5\cdot(m-1)$; the total query cost budget is set to $\Lambda=750$ for every task; and the observation noise variance is set to $\sigma^2=10^{-3}$. The performance of all methods is measured by the simple regret in (53).

\begin{figure}[t]
  \centering
  \centerline{\includegraphics[scale=0.39]{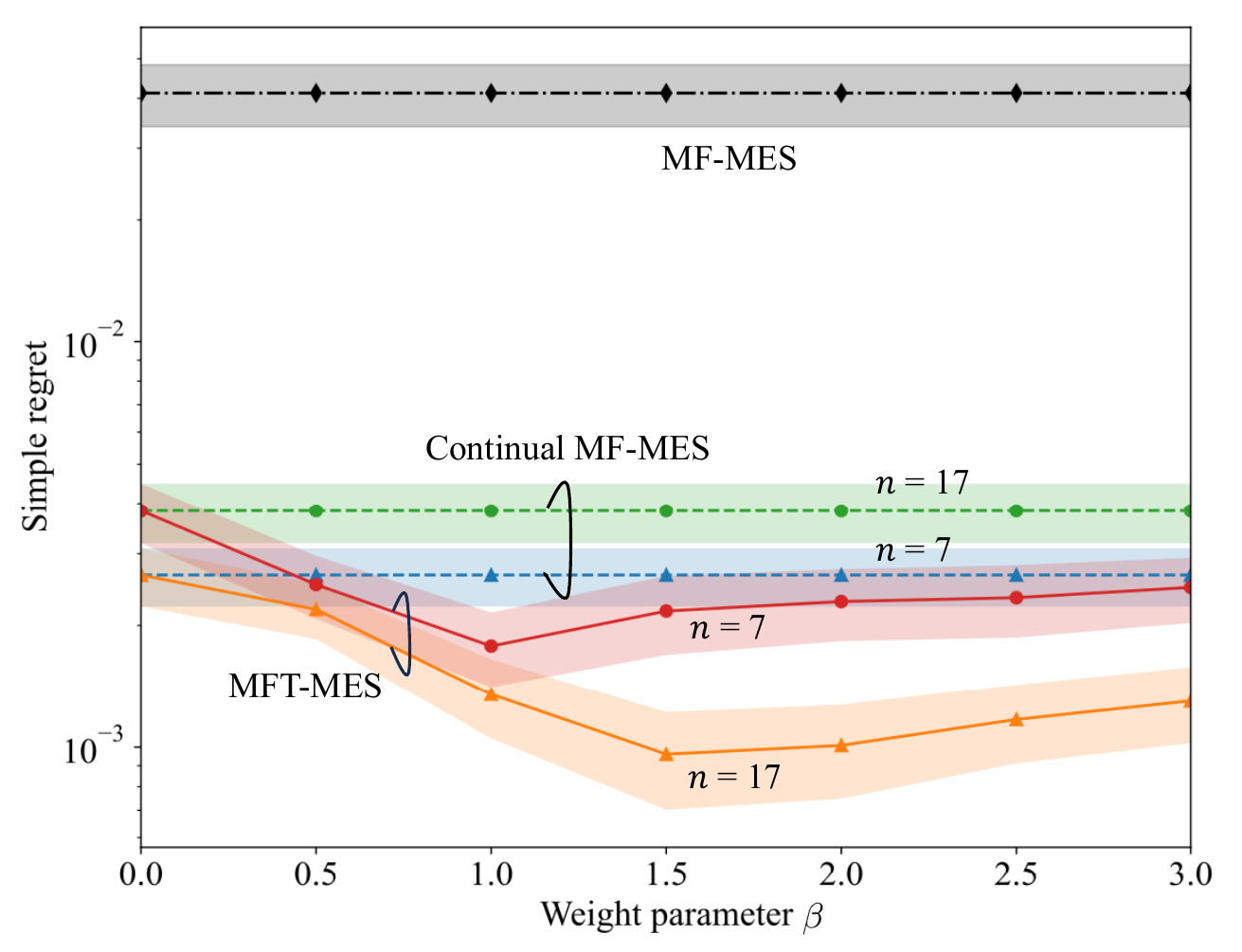}}
  \figuretag{10}
  \caption{Gas emission source term estimation: Simple regret (53) against weight parameter $\beta$, for MF-MES, Continual MF-MES with $n=7$ and $n=17$ tasks observed, and MFT-MES with $n=7$ and $n=17$ tasks observed. The number of particles for Continual MF-MES and MFT-MES is set to $V=10$.}\label{fig: gas sr vs beta}
\end{figure}

In Fig. \ref{fig: gas sr vs n}, we set the weight parameter in (37) to $\beta=1.5$ for MFT-MES, and plot the simple regret (53) as a function of the number of tasks $n$ observed so far. The results demonstrate again the capacity of both Continual MF-MES and MFT-MES to transfer knowledge across tasks, achieving better performance as compared to MF-MES. After processing at least four tasks, MFT-MES outperforms Continual MF-MES. In particular, MFT-MES obtains a lower simple regret by a factor of two as compared to Continual MF-MES at the end of task $n=17$, confirming the importance of accounting for knowledge transfer in the acquisition function (37).

In a manner similar to Sec. VI-D, we demonstrate the impact of weight parameter $\beta$ on the simple regret evaluated at the 7-th and 17-th task in Fig. \ref{fig: gas sr vs beta}. The superiority of MFT-MES over all other schemes is observed to hold for any values of weight parameter $\beta>0$. MFT-MES achieves the best performance with weight parameter around $\beta=1.0$ for $n=7$, and approximately $\beta=1.5$ for $n=17$. The overall trend confirms the discussion in Sec. VI-C, as a larger value of weight parameter $\beta$ is more desirable when the number of tasks $n$ increases.

\bibliographystyle{IEEEtran}

\bibliography{sup_refer}
